\documentclass[11pt]{article}
\pdfoutput=1

\usepackage[final]{acl}

\usepackage{times}
\usepackage{latexsym}

\usepackage[T1]{fontenc}

\usepackage[utf8]{inputenc}

\usepackage{microtype}

\usepackage{inconsolata}

\usepackage{graphicx}
\usepackage{multirow}
\usepackage{booktabs}
\usepackage{xcolor}
\usepackage{makecell}
\usepackage{subcaption}
\usepackage{amsmath}
\usepackage{array}
\usepackage{enumitem}
\usepackage{arydshln}
\usepackage{url}

\title{CARPAS: Towards Content-Aware Refinement of Provided Aspects for Summarization in Large Language Models}

\author{
    Yong-En Tian$^{\dagger}$, Yu-Chien Tang$^{\dagger}$, An-Zi Yen, Wen-Chih Peng \\
    Department of Computer Science, National Yang Ming Chiao Tung University, Taiwan\\
    \texttt{bryanttian.cs12@nycu.edu.tw, tommytyc.cs10@nycu.edu.tw,}\\ \texttt{azyen@nycu.edu.tw, wcpeng@cs.nycu.edu.tw} \\
}

\begin{document}
\maketitle

\def\thefootnote{$\dagger$}\footnotetext{Both authors contributed equally to this work.}\def\thefootnote{\arabic{footnote}}

\begin{abstract}
Aspect-based summarization has attracted significant attention for its ability to generate more fine-grained and user-aligned summaries. 
While most existing approaches assume a set of predefined aspects as input, real-world scenarios often present challenges where these given aspects may be incomplete, irrelevant, or entirely missing from the document. 
Users frequently expect systems to adaptively refine or filter the provided aspects based on the actual content. 
In this paper, we initiate this novel task setting, termed Content-Aware Refinement of Provided Aspects for Summarization (CARPAS), with the aim of dynamically adjusting the provided aspects based on the document context before summarizing.
We construct three new datasets and conduct pilot experiments with four representative prompting strategies. 
Our analysis reveals that LLMs often over-generate aspects, resulting in excessively long and misaligned summaries. 
Building on this observation, we introduce a two-stage framework that derives lightweight scope guidance before aspect refinement and summarization, improving focus and reducing over-generation. 
Our extensive experiments show that the proposed approach significantly improves performance across all datasets.
Moreover, our deeper analyses uncover LLMs' compliance when the requested number of aspects differs from their own estimations, establishing a crucial insight for the deployment of LLMs in similar real-world applications.\footnote{Code and datasets are available at \url{https://github.com/bryant-nn/CARPAS}}
\end{abstract}

\section{Introduction}

Aspect-based summarization (ABS) offers tailored summaries that focus on specific topics or facets within documents, enabling users to quickly extract targeted information. This fine-grained summarization approach has proven highly valuable across multiple domains, such as customer reviews \citep{Summary_product}, scientific papers \citep{ACLSum}, and financial earnings call transcripts \citep{Template-based_Financial}.
While existing ABS methods have shown strong performance, they typically rely on a predefined and accurate set of aspects as input.
These approaches can broadly be categorized into two main types.
The first type adopts a two-stage methodology, where the initial stage is dedicated to generating aspect-based elements, such as identifying key sentences \citep{WikiASP, OpenAsp} or relevant keywords \citep{TWAG, AnyAspect} for each aspect.
In the second stage, this identified and aspect-specific content is then fed into a summarization model to produce the final summary. 
\begin{figure}[t]
    \centering
    \includegraphics[width=\linewidth]{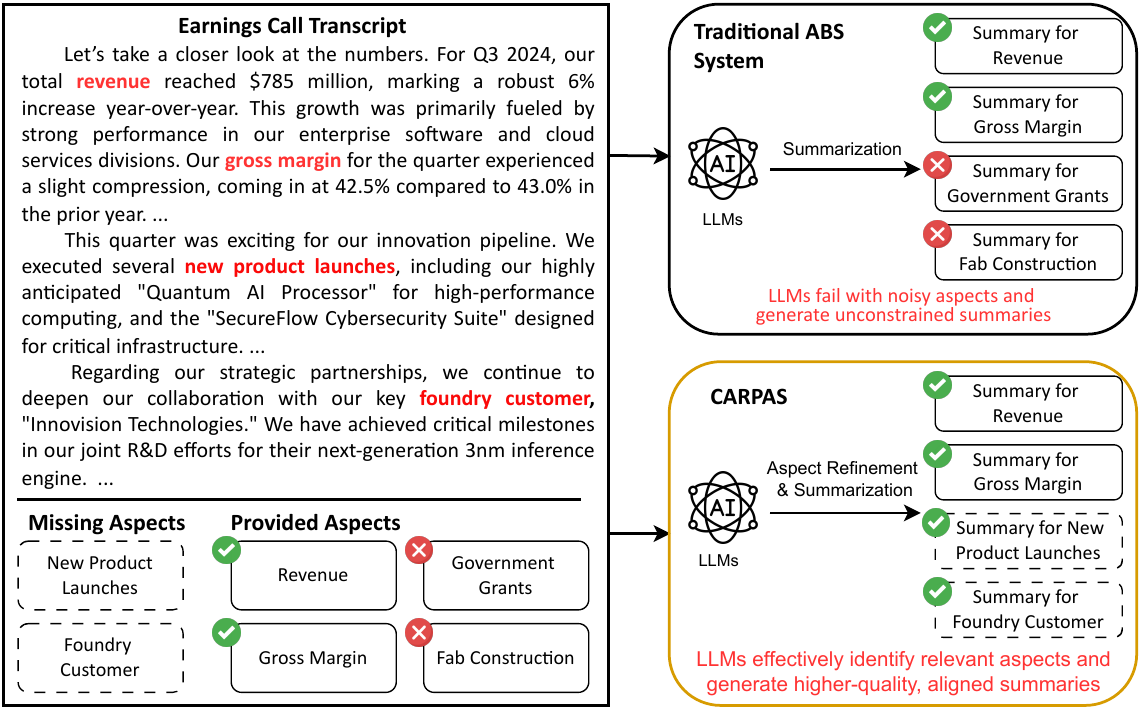}
    \caption{CARPAS aims to refine the provided aspects to generate more aligned summaries, a task that is challenging for previous ABS approaches.} 
    \label{fig:scenario}
\end{figure}
Alternatively, the second type employs an end-to-end approach, where the model directly generates the summary using the entire source document and aspects as input \citep{MODABS, ACLSum}.
However, this inherent reliance on fixed or pre-extracted aspects presents a significant limitation when the given aspects are incomplete, irrelevant, or entirely absent from the document content, a common challenge in real-world applications.
For instance, in the scenario illustrated in Figure \ref{fig:scenario}, the document's ground-truth aspects include \texttt{Revenue}, \texttt{Gross Margin}, \texttt{New Product Launches}, and \texttt{Foundry Customer}. 
Yet, the initially provided aspects might be \texttt{Revenue}, \texttt{Gross Margin}, \texttt{Government Grants}, and \texttt{Fab Construction}. 
Traditional ABS approaches would attempt to generate summaries for all the provided aspects, regardless of their actual relevance. 
Ideally, a system should not only filter out the irrelevant provided aspects (e.g., \texttt{Government Grants} and \texttt{Fab Construction}) but also identify the missing ones that are present in the document (such as \texttt{New Product Launches} and \texttt{Foundry Customer}). 
This gap motivates us to develop a new task: Content-Aware Refinement of Provided Aspects for Summarization (CARPAS).

Given the remarkable advancements of LLMs in recent years \citep{llm_survey, llama3, gpt_4o, deepSeek}, we conduct preliminary experiments to address this challenge using an end-to-end approach.
In this setup, LLMs are tasked with first predicting the correct aspects, and then generating corresponding summaries based on these predicted aspects. 
We explore four representative prompting strategies: direct prompting \citep{few-show-learner}, Chain-of-Thought prompting \citep{CoT}, Chain-of-Thought with self-consistency \citep{CoT_SC}, and Self-Refine \citep{self_refine}. 
However, our initial results indicate that relying solely on the LLMs' inherent capabilities with these prompting strategies achieves an average BERTScore of only 57.36\% on the earnings call transcripts dataset, an unsatisfactory performance for accurate content-aware aspect refinement in real-world deployment. 
In this pilot study, we observe that LLMs tend to predict a more comprehensive set of aspects, often leading to the generation of overly extensive summaries. 
This behavior likely stems from LLMs attempting to cover the entire document by generating an excessive number of aspects.

To mitigate this, we introduce a two-stage framework in which Stage 1 estimates the number of relevant aspects and Stage 2 conditions the LLM's aspect refinement and summarization on this estimate, providing a stronger foundation for inference.
As there is no existing dataset specifically tailored for this task and the predominant cost to annotate one, we leverage LLM to generate synthetic data for two distinct domains: earnings call transcripts and COVID-19 press conference materials. 
Additionally, to assess its real-world applicability, we augment real-world earnings call transcript data to further evaluate the practicality and effectiveness of our subtask.
Our experiments in these three datasets and four prompting strategies demonstrate that our proposed approach significantly improves performance in all experimental settings.

In sum, the main contributions are threefold: 
(1) We propose CARPAS, a practical task for aspect-based summarization, and
release the corresponding code and datasets. 
(2) We empirically validate that prompting LLMs cannot consistently generate great summaries in CARPAS, and thus design a simple yet effective method that can significantly enhance the performance by predicting the number of relevant aspects in advance.
(3) Quantitative results highlight the validity of integrating the aspect-counts prediction into CARPAS.
Our findings further reveal LLMs' unconditional faith in the counts of initially provided aspect sets, offering a reflective view to direct employment of LLMs in similar industrial settings.

\section{Related Work}
\subsection{Aspect-Based Summarization Datasets}

Early ABS datasets assume predefined aspects, often derived from structural cues like Wikipedia section titles \citep{WikiASP}, manually curated labels in scientific domains (e.g., purpose, methodology) \citep{ACLSum, FacetedSum}, or fixed categories for news and meetings \citep{Aspect_news, meeting_abs_dataset_2023}.
More recent work supports open or dynamic aspects to better reflect real-world scenarios where aspects may be implicit or user-defined. 
For instance, OpenAsp \citep{OpenAsp} and OASum \citep{OASum} create open-domain datasets via manual annotation, while other benchmarks use entities as proxy aspects \citep{EntSum} or discover latent aspects in unordered text \citep{disorder_DABS}.
Despite these advances, existing datasets still assume aspects are either explicitly provided or reliably extracted from the document, an assumption that fails in flexible, user-guided scenarios. 
Crucially, no dataset is designed for the CARPAS task, where models must dynamically refine, filter, or reinterpret potentially noisy, incomplete, or irrelevant aspect sets based on the document content. 
This gap motivates our construction of synthetic datasets to pioneer this field.

\vspace{-5pt}
\subsection{Aspect-Based Summarization Methods}
The goal of ABS is to produce summaries targeted to specific aspects or facets of the content. 
These methods can broadly be categorized into two main types: two-stage and end-to-end.
Two-stage approaches first identify aspect-relevant content and then summarize it. 
Early work grouped sentences by aspect \citep{summarize_opinions, WikiASP}, while others identify key sentences or keywords \citep{AnyAspect, TWAG} or cluster aspect-indicative tokens in an unsupervised manner \citep{TokenCluster}. 
The extracted content is then passed to a summarization model.
In contrast, end-to-end methods use a unified framework, conditioning generation on aspect-related signals (e.g., tokens, prompts) to jointly learn relevance and summary formulation \citep{CTRLSum, Aspect_news}. 
For example, some models employ multi-objective training for dynamic aspects \citep{MODABS} or latent structures to bypass the need for labeled data \citep{MA_news}.
Recent research increasingly applies LLMs to ABS tasks, leveraging their powerful generation capabilities.
This includes fine-tuning open-source LLMs like LLaMA2 and Mistral \citep{llm_abs}, prompting models like ChatGPT \citep{chatGPT_QFS_ABS}, and hybrid approaches that combine retrieval with LLM generation to improve factual accuracy \citep{SARESG_2025}.
LLM-driven workflows have also been used in specific domains like summarizing financial transcripts \citep{Template-based_Financial}.
However, all these existing methods, including LLM-based pipelines, rely on a pre-defined and accurate set of input aspects.
Our proposed CARPAS task introduces a new setting that challenges models to dynamically refine imperfect aspect candidates, a scenario unaddressed by prior ABS research.

\begin{figure*}[t]
    \centering
    \includegraphics[width=.95\linewidth]{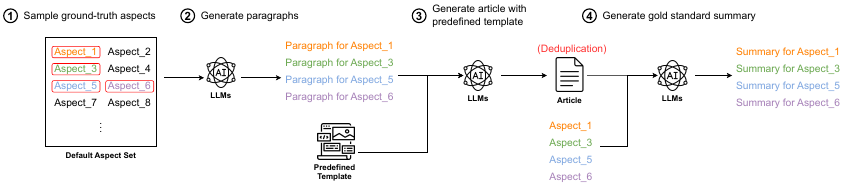}
    \caption{The illustration of synthetic data generation.}
    \label{fig:data_generation}
\end{figure*}

\section{Dataset Construction}

\noindent \textbf{Synthetic data.}
As there is no previous dataset for this task, we construct a synthetic benchmark pipeline for two distinct domains: earnings call transcript (denoted as ECT), and the daily Central Epidemic Command Center COVID-19 press conference (denoted as COVID-19-PC).
These domains are representative of complex real-world scenarios characterized by multiple speakers discussing numerous topics in an interleaved manner.
The dataset construction workflow is in Figure \ref{fig:data_generation}.

\begin{table}[t]
\centering
\small
\setlength{\tabcolsep}{1mm}
\begin{tabular}{p{7.5cm}}
\toprule
\textbf{ECT \& RW-ECT} \\ \hline
\textbf{Profit and Loss Statement Highlights}: Revenue results with QoQ and YoY changes and underlying reasons; revenue guidance; wafer sales including wafer quantity and ASP breakdown; net margin and diluted EPS results. \\

\textbf{Segment or Platform Highlights}: Sales performance by segment or platform, corresponding margin levels, and management commentary; sales guidance, forecasts, or trend outlooks for the next quarter or full year. \\

\textbf{Fab Construction, Expansion and Capacity}: New fab construction progress within the quarter; capacity expansion plans for existing fabs. \\
\midrule
\textbf{COVID-19-PC} \\ \hline
\textbf{Epidemic Statistics Overview}: Daily confirmed cases (local and imported) with DoD or WoW changes and contextual explanation; total deaths and recovered cases with demographic breakdowns when available; testing volumes (e.g., PCR, antigen) and positivity rate trends. \\

\textbf{Vaccination Progress and Plans}: Daily or weekly vaccination counts; coverage rates by age group; updates on booster doses and new vaccine arrivals; commentary on vaccine supply, procurement, and logistics. \\

\textbf{Public Health Policy Updates}: Adjustments to mask mandates, social distancing rules, and gathering limits; border control measures including entry policies and quarantine requirements; changes in public health alert levels. \\
\bottomrule
\end{tabular}
\caption{Example of default aspect sets.}
\label{tab:default_aspect_example}
\end{table}

For ECT, we refer to the aspect set from \citet{Template-based_Financial} which is defined by a financial analyst and comprises 18 key aspects. 
To further enhance the data diversity, we include 21 different company categories.
For COVID-19-PC, we refer to the transcript structure of government COVID-19 press conference\footnote{\url{https://youtube.com/playlist?list=PLSHckwvN6OpIUHJPHeXB1rxqyWhJozfxR}} and define 26 aspects.
We first construct transcript templates based on typical structures observed in real transcripts for both domains, including sections such as opening remarks, discussion, topic highlights, and a Q\&A session, ensuring a realistic document flow.
Second, we randomly sample a total number of aspects (ranging from 4 to 8) from the default aspect set.
Default aspect examples are presented in Table~\ref{tab:default_aspect_example}.
Each sampled aspect will then be rewritten as a paragraph with plausible details and combined into a transcript in line with the transcript templates.
The corresponding aspect-based summary for each aspect will also be generated to serve as the ground-truth of the transcript.
This controlled process ensures that each document is constructed around a precise number of ground-truth aspects and summaries, creating a well-defined benchmark for our experiments.
As a result, a total of 1,163 documents (\~50 for each company category) are synthesized for ECT and 160 for COVID-19-PC.

\begin{table}[t]
\centering
\small
\resizebox{\linewidth}{!}{
\begin{tabular}{lccccccc}
    \toprule
    \multirow{2}{*}{\textbf{Dataset}} & \multicolumn{5}{c}{\textbf{\#Aspects Per Doc.}} & \multicolumn{2}{c}{\textbf{Avg. \#Tokens}} \\
    \cmidrule(lr){2-6} \cmidrule(lr){7-8} & \textbf{4} & \textbf{5} & \textbf{6} & \textbf{7} & \textbf{8} & \textbf{\makecell[l]{Aspect \\Summary}} & \textbf{Doc.} \\
    \midrule
    ECT & 231 & 234 & 231 & 233 & 234 & 103.33 & 4532.67 \\
    COVID-19-PC & 31 & 34 & 33 & 31 & 31 & 86.43 & 2299.40 \\
    RW-ECT     & 19 & 17 & 20 & 20 & 17 & 105.54 & 3745.00 \\
    \bottomrule
\end{tabular}%
}
\caption{Dataset statistics grouped by number of aspects and the average number of tokens in each document.}
\label{tab:aspect-stats}
\end{table}

\noindent \textbf{Real-world data.}
We collect earnings call transcripts from five leading semiconductor manufacturing companies, spanning from 2019 to 2024, as reference sources for the detailed content paragraphs for each sampled aspect. 
This augmentation process results in 93 real-world ECT documents (denoted as RW-ECT). 

Table \ref{tab:aspect-stats} summarizes the key statistics of all datasets, including document counts, aspect numbers, document tokens, and aspect summary tokens.

\begin{figure*}[t]
    \centering
    \includegraphics[width=.9\linewidth]{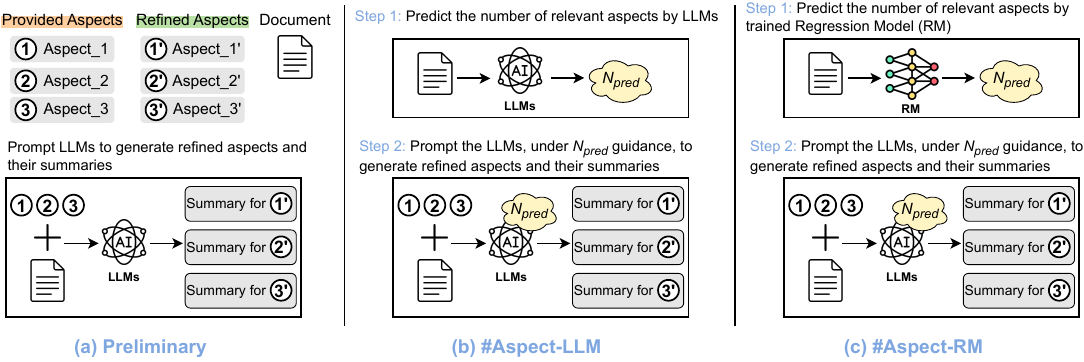}
    \caption{Overall workflows of three approaches for CARPAS. The Preliminary method (a) directly refines the provided aspects using prompting strategies. The \#Aspect-LLM method (b) adds aspect number prediction via LLMs. The \#Aspect-RM method (c) incorporates aspect number prediction from a trained regression model.}
    \label{fig:method}
\end{figure*}

\section{Method}

\subsection{Prompting Strategies}
In the Preliminary Experiments, as stated in Figure \ref{fig:method} (a),  we employ four representative prompting strategies to tackle the CARPAS task: direct prompting \citep{few-show-learner}, Chain-of-Thought prompting \citep{CoT}, Chain-of-Thought with self-consistency \citep{CoT_SC}, and Self-Refine \citep{self_refine}. 
These strategies are utilized to refine the provided aspects and subsequently generate the corresponding aspect-based summaries.
For each strategy, the LLMs are given the full document and the provided aspects (which could include irrelevant or missing ones). 
The goal is to produce refined versions of the provided aspects that are relevant to the document, followed by a summary for each refined aspect.
The four strategies are defined as follows:

\noindent \textbf{Direct Prompting (DP)}: This approach involves providing the LLM with a straightforward instruction that describes the task, requesting the refined aspects and summaries directly as the final output.

\noindent \textbf{Chain-of-Thought Prompting (CoT)}: This strategy prompts the LLM to first generate a step-by-step explanation of its reasoning for refining the provided aspects (i.e., why specific aspects are retained, modified, removed, or added) before generating the final answer.

\noindent \textbf{Chain-of-Thought with Self-Consistency (CoT-SC)}: As an extension of CoT, this approach generates multiple, diverse reasoning paths for the same task. 
The final, most consistent answer is then determined by a concluding LLM prompt that instructs the model to evaluate all generated paths and select the most coherent and logical one, which enhances the robustness of the results.

\noindent \textbf{Self-Refine (SR)}: This approach adopts an iterative, agentic framework with a conditional workflow.
The first agent evaluates the provided aspects to determine whether refinement is necessary.
If refinement is required, the second agent performs the necessary modifications and then returns the revised aspects to the first agent for re-evaluation.
This process may repeat until the first agent determines that no further refinement is needed.
Once the aspects are finalized, the third agent generates summaries based on them.

\subsection{Aspect Number Prediction}
During the Preliminary Experiments, we observe that LLMs often struggle to generate summaries that accurately reflect the document content.
As shown in Table \ref{tab:combined_multi_metric_summary} and Figure \ref{fig:ECT-asp}, the final summaries achieve unsatisfactory BERTScore and ROUGE-L (approximately 60\% and 30\%), primarily due to a mismatch between the total number of predicted aspects and the ground-truth annotations.
To address this limitation and guide the LLMs more effectively, we introduce a dedicated subtask: Aspect Number Prediction. 
We hypothesize that the suboptimal performance observed in preliminary experiments stems from the LLMs' tendency to comply with the total number of provided aspects, and with proper guidance, such as prompting the model with an accurate estimate of the number of aspects, LLMs are capable of correctly predicting the aspect set.
For this subtask, the input is the full document, and the output is a single integer representing the estimated number of relevant aspects within that document. 
We explore two approaches for this prediction, and the overall workflows are shown in Figure \ref{fig:method} (b),~(c).
\subsubsection{\#Aspect-LLM}
The first approach involves directly prompting LLMs to predict the number of relevant aspects based on the document content by asking them ``Estimate how many distinct, non-overlapping aspects or key themes are present''. 
This method aims to test whether LLMs themselves can precisely predict the aspect number.

\subsubsection{\#Aspect-RM}
The second approach focuses on training a regression model composed of an embedding model followed by a linear layer to predict the correct aspect number.
Specifically, we use \texttt{Qwen3-Embedding-0.6B} \citep{qwe3_embedding} as the embedding model and train the regression model by minimizing the Mean Absolute Error (MAE) loss.
We perform this fine-tuning on the training split of our synthetic datasets, where the ground-truth aspect number for each document is known from its generative process.

\subsubsection{Combined with Aspect Refinement}
The predicted aspect number $N_{pred}$ from either of these methods is then applied to the four prompting strategies, serving as a guidance for the LLMs in the subsequent aspect refinement and summarization steps. 
LLMs are expected to output refined aspects by tasking them with following prompt: ``This document likely contains around $N_{pred}$ aspects. Please identify the $N_{pred}$ most relevant aspects and provide a concise summary for each.''
This allows for precise control over the output granularity, and can further prevent the model from generating overly detailed or sparse aspect sets, an issue that often arises when the LLMs misinterpret the correct aspects.

\section{Experiments and Discussion}

\begin{table*}[t]
\centering
\resizebox{\textwidth}{!}{%
\setlength{\tabcolsep}{1.5mm} 

\begin{tabular}{l *{24}{c}}
\toprule
& \multicolumn{8}{c}{\textbf{ECT}} & \multicolumn{8}{c}{\textbf{COVID-19-PC}} & \multicolumn{8}{c}{\textbf{RW-ECT}}\\
\cmidrule(lr){2-9} \cmidrule(lr){10-17} \cmidrule(lr){18-25}
\multirow{2}{*}{\makecell[c]{\textbf{Prompting}\\\textbf{Strategy}}}
& \multicolumn{2}{c}{\textbf{G3-12B}} & \multicolumn{2}{c}{\textbf{G3-27B}} & \multicolumn{2}{c}{\textbf{4o-mini}} & \multicolumn{2}{c}{\textbf{4o}}
& \multicolumn{2}{c}{\textbf{G3-12B}} & \multicolumn{2}{c}{\textbf{G3-27B}} & \multicolumn{2}{c}{\textbf{4o-mini}} & \multicolumn{2}{c}{\textbf{4o}} & \multicolumn{2}{c}{\textbf{G3-12B}} & \multicolumn{2}{c}{\textbf{G3-27B}} & \multicolumn{2}{c}{\textbf{4o-mini}} & \multicolumn{2}{c}{\textbf{4o}} \\
\cmidrule(lr){2-3} \cmidrule(lr){4-5} \cmidrule(lr){6-7} \cmidrule(lr){8-9}
\cmidrule(lr){10-11} \cmidrule(lr){12-13} \cmidrule(lr){14-15} \cmidrule(lr){16-17}
\cmidrule(lr){18-19} \cmidrule(lr){20-21} \cmidrule(lr){22-23} \cmidrule(lr){24-25}
& \textbf{BS} & \textbf{R-L} & \textbf{BS} & \textbf{R-L} & \textbf{BS} & \textbf{R-L} & \textbf{BS} & \textbf{R-L} & \textbf{BS} & \textbf{R-L} & \textbf{BS} & \textbf{R-L} & \textbf{BS} & \textbf{R-L} & \textbf{BS} & \textbf{R-L} & \textbf{BS} & \textbf{R-L} & \textbf{BS} & \textbf{R-L} & \textbf{BS} & \textbf{R-L} & \textbf{BS} & \textbf{R-L} \\
\hline
\hline

\multicolumn{25}{c}{\textbf{Preliminary}} \\
\hline 
DP      & 56.1 & 32.6 & 62.6 & 33.6 & 50.9 & 23.7 & 62.7 & \multicolumn{1}{c|}{31.7} & 56.9 & 33.8 & 64.5 & 35.4 & 47.9 & 20.9 & 57.6 & \multicolumn{1}{c|}{27.8} & 54.9 & 29.1 & 62.0 & 30.1 & 48.3 & 19.8 & 61.0 & 27.9\\
CoT       & 58.1 & 32.8 & 63.5 & 34.3 & 45.3 & 21.0 & 61.6 & \multicolumn{1}{c|}{28.3} & 55.2 & 31.8 & 64.2 & 35.3 & 40.9 & 17.6 & 60.0 & \multicolumn{1}{c|}{27.0} & 54.6 & 29.5 & 62.3 & 31.6 & 43.0 & 18.3 & 59.8 & 25.6 \\
CoT-SC    & 63.6 & 34.1 & 63.1 & 31.1 & 48.5 & 23.4 & 63.7 & \multicolumn{1}{c|}{30.6} & 66.2 & 38.1 & 65.2 & 34.7 & 46.2 & 21.4 & 62.2 & \multicolumn{1}{c|}{30.4} & 61.2 & 30.6 & 62.1 & 29.0 & 46.8 & 20.2 & 62.5 & 28.2 \\
SR & 56.6 & 32.2 & 51.7 & 28.3 & 53.2 & 25.7 & 56.6 & \multicolumn{1}{c|}{26.6} & 60.2 & 35.3 & 52.2 & 29.6 & 52.7 & 26.7 & 56.0 & \multicolumn{1}{c|}{28.9} & 55.4 & 29.2 & 51.4 & 26.6 & 51.4 & 24.0 & 54.7 & 25.9\\
\hline 
\hline 
\multicolumn{25}{c}{\textbf{\#Aspect-LLM}} \\
\hline 
DP      & 53.5 & 30.7 & 61.6 & 32.0 & 50.9 & 23.0 & 56.3 & \multicolumn{1}{c|}{27.7} & 54.2 & 32.2 & 61.0 & 33.2 & 55.1 & 23.6 & 56.2 & \multicolumn{1}{c|}{26.5} & 56.0 & 29.1 & 59.0 & 27.9 & 51.1 & 20.7 & 58.7 & 25.7 \\
CoT       & 59.5 & 32.8 & 62.8 & 32.6 & 52.3 & 23.5 & 55.8 & \multicolumn{1}{c|}{25.1} & 58.8 & 34.0 & 61.4 & 33.6 & 53.4 & 22.5 & 55.8 & \multicolumn{1}{c|}{24.5} & 60.5 & 31.1 & 60.4 & 29.7 & 52.9 & 21.6 & 58.2 & 24.1 \\
CoT-SC    & 59.2 & 31.5 & 60.7 & 29.6 & 56.1 & 26.0 & 60.8 & \multicolumn{1}{c|}{28.6} & 55.6 & 31.7 & 59.9 & 31.2 & 55.3 & 24.5 & 56.1 & \multicolumn{1}{c|}{26.8} & 56.3 & 26.6 & 57.7 & 25.7 & 51.6 & 21.6 & 58.3 & 25.3 \\
SR & 56.1 & 32.0 & 53.2 & 29.0 & 51.7 & 26.6 & 55.3 & \multicolumn{1}{c|}{28.7} & 54.0 & 30.8 & 60.4 & 34.3 & 54.0 & 26.3 & 57.0 & \multicolumn{1}{c|}{28.1} & 45.4 & 21.9 & 58.4 & 29.3 & 51.0 & 22.5 & 59.1 & 26.2 \\
\hline 
\hline 
\multicolumn{25}{c}{\textbf{\#Aspect-RM}} \\
\hline 
DP      & 70.7 & \textbf{38.6} & \underline{72.9} & 36.7 & 59.7 & 26.7 & \underline{71.5} & \multicolumn{1}{c|}{\underline{34.0}} & 69.7 & \underline{39.8} & \textbf{74.4} & 38.9 & 64.2 & 26.5 & 71.0 & \multicolumn{1}{c|}{32.4} & \underline{68.9} & \textbf{34.7} & \underline{73.0} & 33.9 & 59.4 & 23.6 & \underline{71.4} & \underline{31.2}\\
CoT       & \underline{71.2} & \underline{38.3} & \textbf{73.5} & \textbf{37.7} & 61.4 & 27.3 & 69.3 & \multicolumn{1}{c|}{30.3} & \underline{70.1} & 39.4 & \underline{74.2} & \underline{39.4} & 61.6 & 25.4 & 69.0 & \multicolumn{1}{c|}{30.1} & 68.0 & \underline{34.0} & \textbf{73.1} & \textbf{34.9} & 59.5 & 23.8 & 68.1 & 27.7 \\
CoT-SC    & \textbf{72.0} & 35.7 & 71.9 & 33.6 & \underline{69.9} & \underline{31.2} & 71.0 & \multicolumn{1}{c|}{32.3} & \textbf{73.9} & \textbf{41.1} & 74.0 & 37.2 & \underline{68.7} & \underline{29.5} & \underline{72.4} & \multicolumn{1}{c|}{\underline{33.7}} & \textbf{71.5} & 32.6 & 71.8 & 31.4 & \underline{69.2} & \underline{27.8} & 71.2 & 30.1 \\
SR & 65.2 & 34.2 & 70.6 & \underline{37.0} & \textbf{70.6} & \textbf{33.5} & \textbf{71.9} & \multicolumn{1}{c|}{\textbf{34.4}} & 66.9 & 38.1 & 72.4 & \textbf{40.3} & \textbf{71.2} & \textbf{34.1} & \textbf{73.1} & \multicolumn{1}{c|}{\textbf{35.9}} & 63.0 & 30.4 & 70.3 & \underline{34.7} & \textbf{70.2} & \textbf{30.7} & \textbf{71.7} & \textbf{31.7} \\
\hline\hline 

\end{tabular}%
}
\caption{Results ($\times 100$\%) on synthetic (ECT and COVID-19-PC) and real-world (RW-ECT) datasets, with BS standing for BERTScore, and R-L standing for ROUGE-L. The highest results for each metric are in \textbf{boldface}, while the second-best are \underline{underlined}.}
\label{tab:combined_multi_metric_summary}
\end{table*}
To systematically investigate the aspect number prediction 
and LLM behavior in CARPAS, we address the following research questions:

\noindent \textbf{RQ1:} Can the predicted number of aspects accurately match the ground-truth and human preference? (Section~\ref{sec:quantitative}) 

\noindent \textbf{RQ2:} How does the accuracy of aspect count prediction affect the quality of generated aspects and summaries? (Section~\ref{sec:impact_number})  

\noindent \textbf{RQ3:} When the predicted count is inaccurate, can LLMs adjust, or do they exhibit compliance behavior? (Section~\ref{sec:sensitivity})

\subsection{Experimental Setup. }
\noindent \textbf{LLMs and Prompting Strategies}
In our experiments, we utilize Gemma-3 12B (G3-12B), Gemma-3 27B (G3-27B) \citep{gemma_3}, GPT-4o-mini (4o-mini), and GPT-4o (4o) \citep{gpt_4o} as the LLMs. 
For all LLM interactions, the temperature is set to 0.7 to balance creativity and consistency in generation. 
The maximum number of iterations for Self-Refine is set to 16.

\noindent \textbf{Datasets. }
The train-test splitting is 9:1.
As for the provided aspects, we simulate real-world scenarios with three different settings, where $y$, $n$ denote the correct and incorrect number of aspects: 
(1) Completely incorrect aspects: $(y, n) \in \{(0,2), (0,4),(0,6)\}$.
(2) Completely correct aspects: $(y, n) \in \{(2,0), (4,0),(6,0)\}$.
(3) A mix of correct and incorrect aspects: $(y, n) \in \{(2,2), (4,4),(6,6)\}$.
For simplicity, we adopt a shorthand notation in the form of \texttt{y2n2} to represent $(y, n) = (2,2)$ in the following sections.

\noindent \textbf{Implementation Details of \#Aspect-RM. }
The Regression Model is trained using the training splits of each dataset.
Specifically, we fine-tune the pre-trained embedding model Qwen3-Embedding-0.6B to perform regression on the number of aspects associated with a given document. 
To adapt the model for this task, we append a lightweight regression head on top of the document embedding derived from the base model.
The document representation is obtained by applying mean pooling over the final hidden states of the encoder. 
This pooled embedding is passed through a two-layer feed-forward regression head composed of a linear transformation that reduces the hidden size by half, followed by a ReLU activation, a dropout layer with a rate of 0.2, and a final linear projection to a single scalar output.
Through our experiments, we apply LoRA \citep{lora} for fine-tuning, which effectively reduces memory consumption and enables more efficient model training, and we use AdamW \citep{adamw} as the optimizer.
The model is trained for 30 epochs with a batch size of 1, a learning rate of 2e-5, and a fixed random seed of 42. 

\noindent \textbf{Evaluation Metrics. }
To measure the quality of the generated summary, we use BERTScore \citep{BERTScore} along with ROUGE-1, ROUGE-2, and ROUGE-L \citep{rouge} as the main evaluation metrics, following \citep{disorder_DABS}.
The BERT-based model for BERTScore is \texttt{deberta-v3-large} \citep{deberta-v3}.
In addition, we assess the accuracy of aspect identification by calculating the absolute difference in the number of aspects between the reference and generated summaries (denoted as \#AbsAspDiff, the lower the better).
More detailed evaluation metrics and results are described in the Appendix \ref{app:full_results}.

\begin{figure*}[t]
    \centering
    \begin{subfigure}[b]{0.3\textwidth}
        \includegraphics[width=\linewidth]{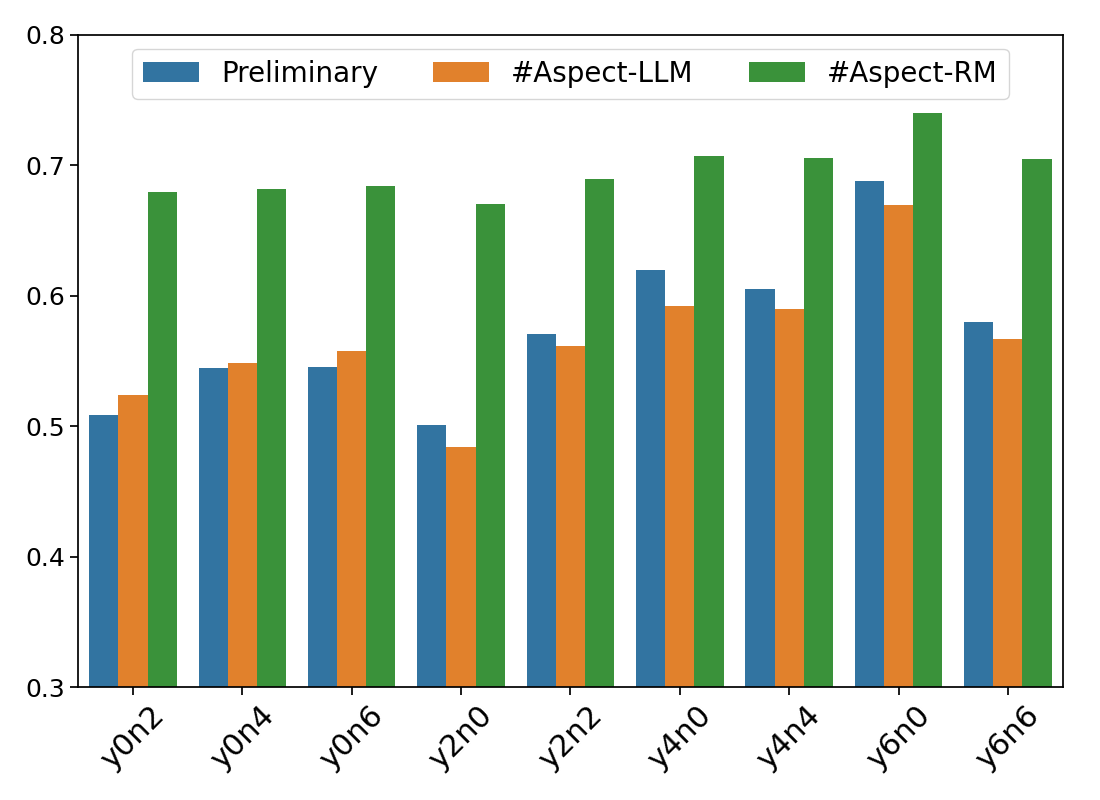}
        \caption{ECT}
    \end{subfigure}
    \hfill
    \begin{subfigure}[b]{0.3\textwidth}
        \includegraphics[width=\linewidth]{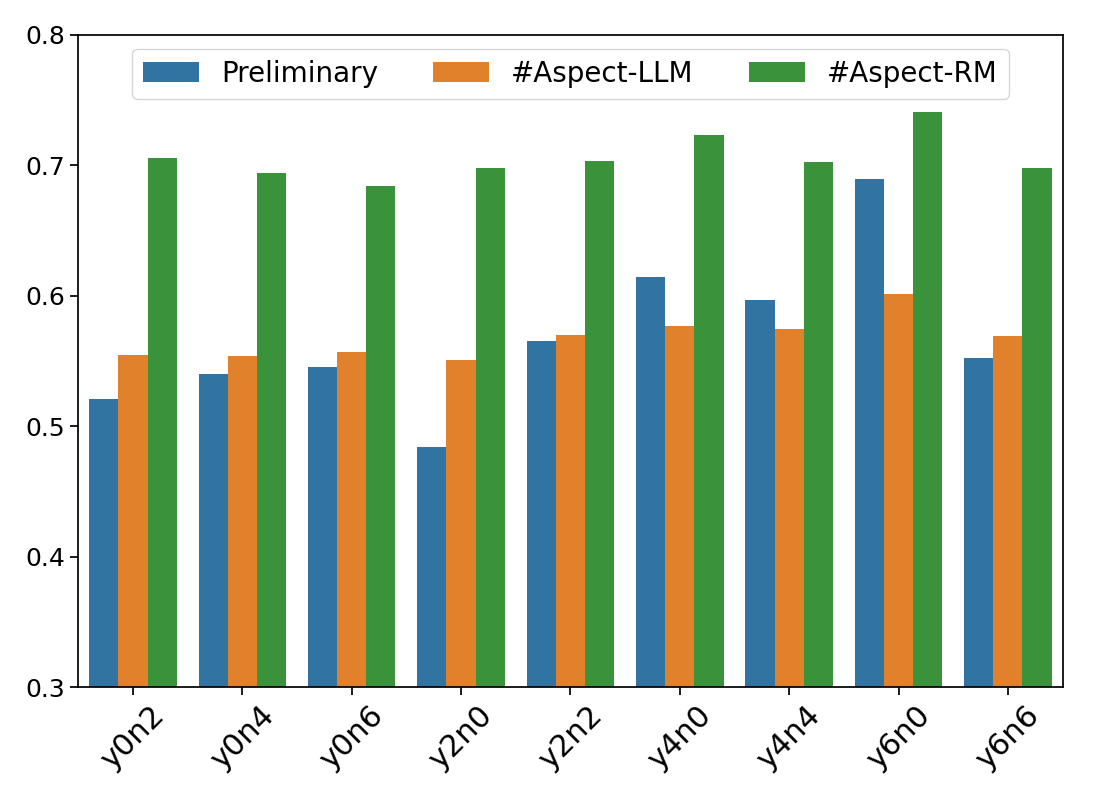}
        \caption{COVID-19-PC}
    \end{subfigure}
    \hfill
    \begin{subfigure}[b]{0.3\textwidth}
        \includegraphics[width=\linewidth]{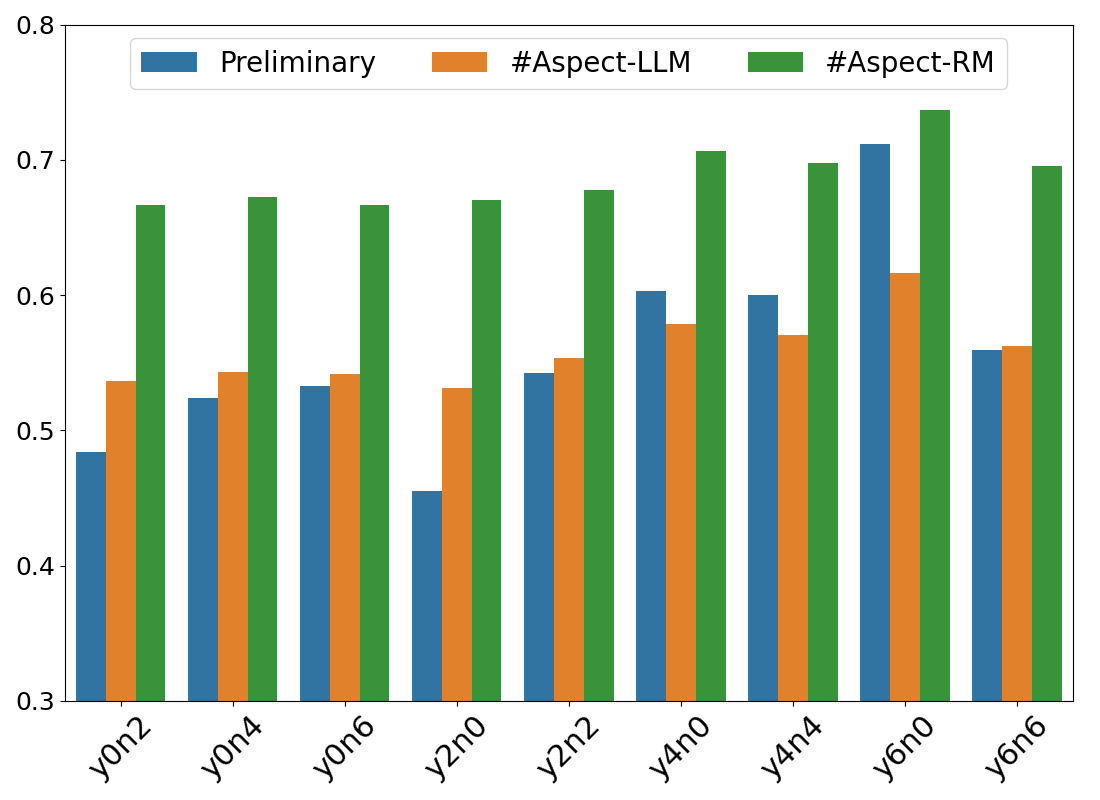}
        \caption{RW-ECT}
    \end{subfigure}
    \caption{Average BERTScore results on each dataset across different provided aspect settings.}
    \label{bert-setting}
\end{figure*}

\subsection{RQ1: Quantitative Results}\label{sec:quantitative}
Table \ref{tab:combined_multi_metric_summary} presents the overall performance of all LLMs and prompting strategies across synthetic (ECT and COVID-19-PC) and real-world (RW-ECT) datasets. 
It can be seen that \#Aspect-RM significantly improves both BERTScore and ROUGE-L within Preliminary Experiments settings, and consistently outperforms \#Aspect-LLM.
As shown in Figure \ref{bert-setting}, this superior performance is consistent across all provided aspect settings under different datasets.
Compared to the baseline in Preliminary Experiments, \#Aspect-RM achieves an average improvement of 25.4\% in BERTScore and 19.8\% in ROUGE-L on the ECT dataset. 
Even larger gains are observed on the COVID-19-PC dataset, with BERTScore and ROUGE-L increasing by 30.1\% and 24.4\%, respectively.
This highlights the impact of providing accurate numerical guidance as a prerequisite for generating high-quality, aspect-aligned summaries.
On RW-ECT, \#Aspect-RM continues to deliver substantial gains, with BERTScore and ROUGE-L increasing by 23.4\% and 15.9\%, demonstrating that our proposed subtask is not only effective on synthetic data but also robustly enhances summarization quality on complex, real-world financial documents.

\begin{figure}[t]
    \centering
    \includegraphics[width=\linewidth]{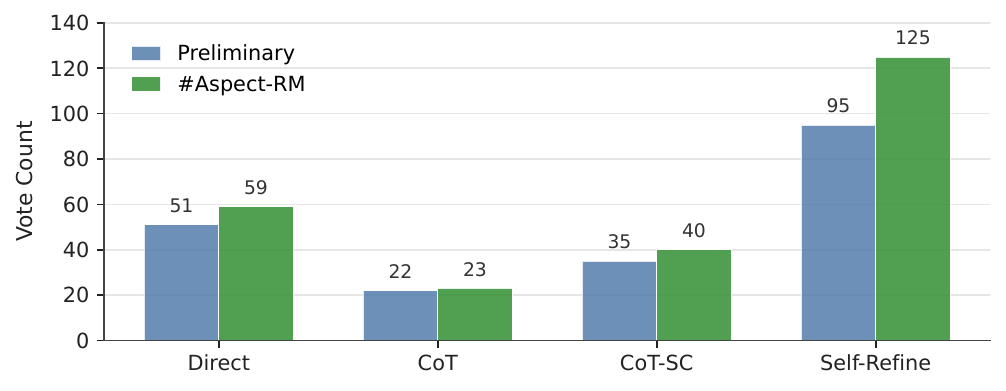}
    \caption{Human preference voting counts on Preliminary and \#Aspect-RM.}
    \label{fig:human_votes}
\end{figure}

To analyze whether the generated aspects and summaries are more human-aligned, we conduct a human preference voting.
Specifically, we invite 12 human annotators to select their preferred aspects and summaries from \#Aspect-RM and Preliminary across four prompting strategies on GPT-4o, and collect a total of 450 preference votes.
The results are illustrated in Figure \ref{fig:human_votes}, and we can observe that \#Aspect-RM is uniformly preferred across different prompting strategies.
This indicates that LLMs can actually generate aspects and summaries aligned with human preference once provided with an accurate estimation of the aspect counts.
Annotation details are provided at Appendix \ref{app:annotation}.

\begin{table*}[ht]
\centering
\resizebox{\textwidth}{!}{%
\begin{tabular}{ll ccc ccc ccc ccc ccc ccc}
\toprule
\multirow{2}{*}{\textbf{Dataset}} & \multirow{2}{*}{\textbf{Method}} & \multicolumn{3}{c}{\textbf{Gemini-2.5-Flash}} & \multicolumn{3}{c}{\textbf{Gemma-3-12B}} & \multicolumn{3}{c}{\textbf{Gemma-3-27B}} & \multicolumn{3}{c}{\textbf{GPT-4o-mini}} & \multicolumn{3}{c}{\textbf{GPT-4o}} & \multicolumn{3}{c}{\textbf{O3-mini}} \\
\cmidrule(r){3-5} \cmidrule(r){6-8} \cmidrule(r){9-11} \cmidrule(r){12-14} \cmidrule(r){15-17} \cmidrule(r){18-20}
& & \textbf{Fact.} & \textbf{Rel.} & \textbf{BS$_{\text{w}}$} & \textbf{Fact.} & \textbf{Rel.} & \textbf{BS$_{\text{w}}$} & \textbf{Fact.} & \textbf{Rel.} & \textbf{BS$_{\text{w}}$} & \textbf{Fact.} & \textbf{Rel.} & \textbf{BS$_{\text{w}}$} & \textbf{Fact.} & \textbf{Rel.} & \textbf{BS$_{\text{w}}$} & \textbf{Fact.} & \textbf{Rel.} & \textbf{BS$_{\text{w}}$} \\
\midrule
\multirow{3}{*}{ECT} & Preliminary & \textbf{4.67} & 4.10 & 0.46 & 4.66 & 3.82 & 0.47 & \textbf{4.69} & 3.82 & 0.43 & 4.62 & 3.32 & 0.43 & 4.71 & 4.04 & 0.47 & \textbf{4.79} & 3.80 & 0.45 \\
& \#Aspect-LLM & \textbf{4.67} & 4.00 & 0.38 & 4.66 & \textbf{3.89} & 0.43 & \textbf{4.69} & 3.78 & 0.43 & 4.62 & 3.64 & 0.46 & \textbf{4.72} & 3.85 & 0.46 & 4.78 & 4.13 & 0.48 \\
& \#Aspect-RM & 4.65 & \textbf{4.22} & \textbf{0.54} & \textbf{4.67} & 3.87 & \textbf{0.50} & 4.68 & \textbf{4.01} & \textbf{0.49} & \textbf{4.67} & \textbf{3.83} & \textbf{0.52} & 4.71 & \textbf{4.12} & \textbf{0.55} & \textbf{4.79} & \textbf{4.19} & \textbf{0.54} \\
\midrule
\multirow{3}{*}{COVID-19-PC} & Preliminary & 4.97 & 4.10 & 0.51 & \textbf{4.99} & 4.16 & 0.54 & 4.98 & 4.19 & 0.50 & 4.88 & 3.49 & 0.51 & \textbf{5.00} & 4.19 & 0.55 & 4.99 & 4.04 & 0.52 \\
& \#Aspect-LLM & \textbf{5.00} & 4.07 & 0.42 & \textbf{4.99} & 4.17 & 0.48 & \textbf{4.99} & 4.29 & 0.47 & 4.97 & 4.06 & 0.53 & 4.99 & 4.16 & 0.50 & 4.99 & 4.37 & 0.55 \\
& \#Aspect-RM & \textbf{5.00} & \textbf{4.47} & \textbf{0.61} & \textbf{4.99} & \textbf{4.38} & \textbf{0.61} & \textbf{4.99} & \textbf{4.48} & \textbf{0.57} & \textbf{4.98} & \textbf{4.23} & \textbf{0.65} & 4.99 & \textbf{4.44} & \textbf{0.64} & \textbf{5.00} & \textbf{4.52} & \textbf{0.62} \\
\midrule
\multirow{3}{*}{RW-ECT} & Preliminary & 3.75 & 3.89 & 0.45 & \textbf{3.79} & \textbf{3.58} & 0.47 & \textbf{3.89} & 3.54 & 0.42 & \textbf{3.98} & 3.08 & 0.45 & 3.89 & 3.80 & 0.48 & \textbf{4.08} & 3.56 & 0.44 \\
& \#Aspect-LLM & 3.65 & 3.88 & 0.36 & 3.76 & 3.57 & 0.39 & 3.83 & 3.68 & 0.40 & 3.83 & 3.54 & 0.42 & 3.89 & 3.71 & 0.45 & 4.01 & 3.83 & 0.45 \\
& \#Aspect-RM & \textbf{3.76} & \textbf{4.01} & \textbf{0.54} & \textbf{3.79} & \textbf{3.58} & \textbf{0.49} & \textbf{3.89} & \textbf{3.72} & \textbf{0.49} & 3.93 & \textbf{3.67} & \textbf{0.52} & \textbf{3.90} & \textbf{3.92} & \textbf{0.54} & 3.98 & \textbf{3.96} & \textbf{0.54} \\
\bottomrule
\end{tabular}%
}
\caption{The factuality (Fact.), relevance (Rel.), and BS$_{\text{w}}$ of predicted aspects across three datasets and models.}
\label{tab:combined_aspect_quality}
\end{table*}

\subsection{RQ2: Impact of Predicted Aspect Number}\label{sec:impact_number}
To delve into the quality of the predicted aspect set, we further evaluate the factuality and relevance between the predicted aspects and the document content by prompting Qwen3-Next-80B-A3B-Instruct on a 1--5 Likert scale.
In addition, we assess the similarity between the predicted and ground-truth aspects using a weighted BERTScore: 
    $BS_{w} = \frac{1}{n+|n-m|}\sum_{i=1}^{n}BS(a^{i}_{\text{pred}},a^{i}_{\text{ground}}),$
where $m$ and $n$ denote the number of predicted and ground-truth aspects, respectively, and $a^{i}_{\text{pred}}$ and $a^{i}_{\text{ground}}$ are the $i$-th aspects in a matched pair.
This metric penalizes the over- or under-estimation of aspect counts and can thus provide a more accurate assessment of the aspect set quality for each method.
For the paired computation of BERTScore, we apply the Hungarian Algorithm \citep{kuhn1955hungarian} to find the optimal pairing of aspects that maximizes the total similarity score.
The results, averaged over four prompting strategies, are presented in Table \ref{tab:combined_aspect_quality}.
Overall, we observe that \#Aspect-RM achieves superior factuality, relevance, and BS$_{\text{w}}$ across all datasets and models, indicating that providing an accurate aspect count estimation to LLMs can effectively improve the quality of the predicted aspects, hence benefiting the downstream ABS task.

\begin{figure}[t] 
    \centering 

    \begin{subfigure}{0.157\textwidth}
        \centering
        \includegraphics[width=\linewidth]{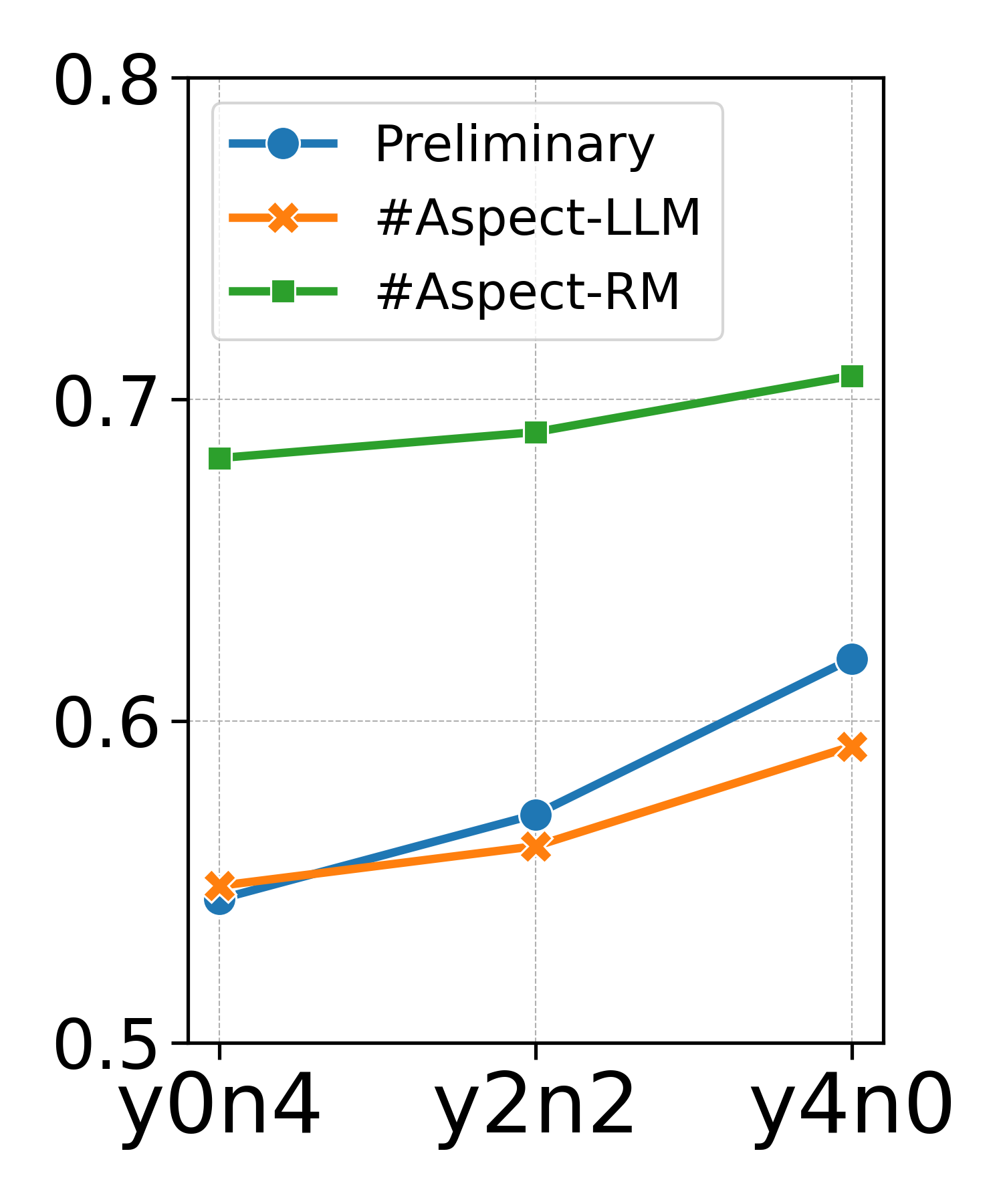}
        \caption{ECT}
        \label{fig:sub1}
    \end{subfigure}%
    \hfill %
    \begin{subfigure}{0.157\textwidth}
         \centering
        \includegraphics[width=\linewidth]{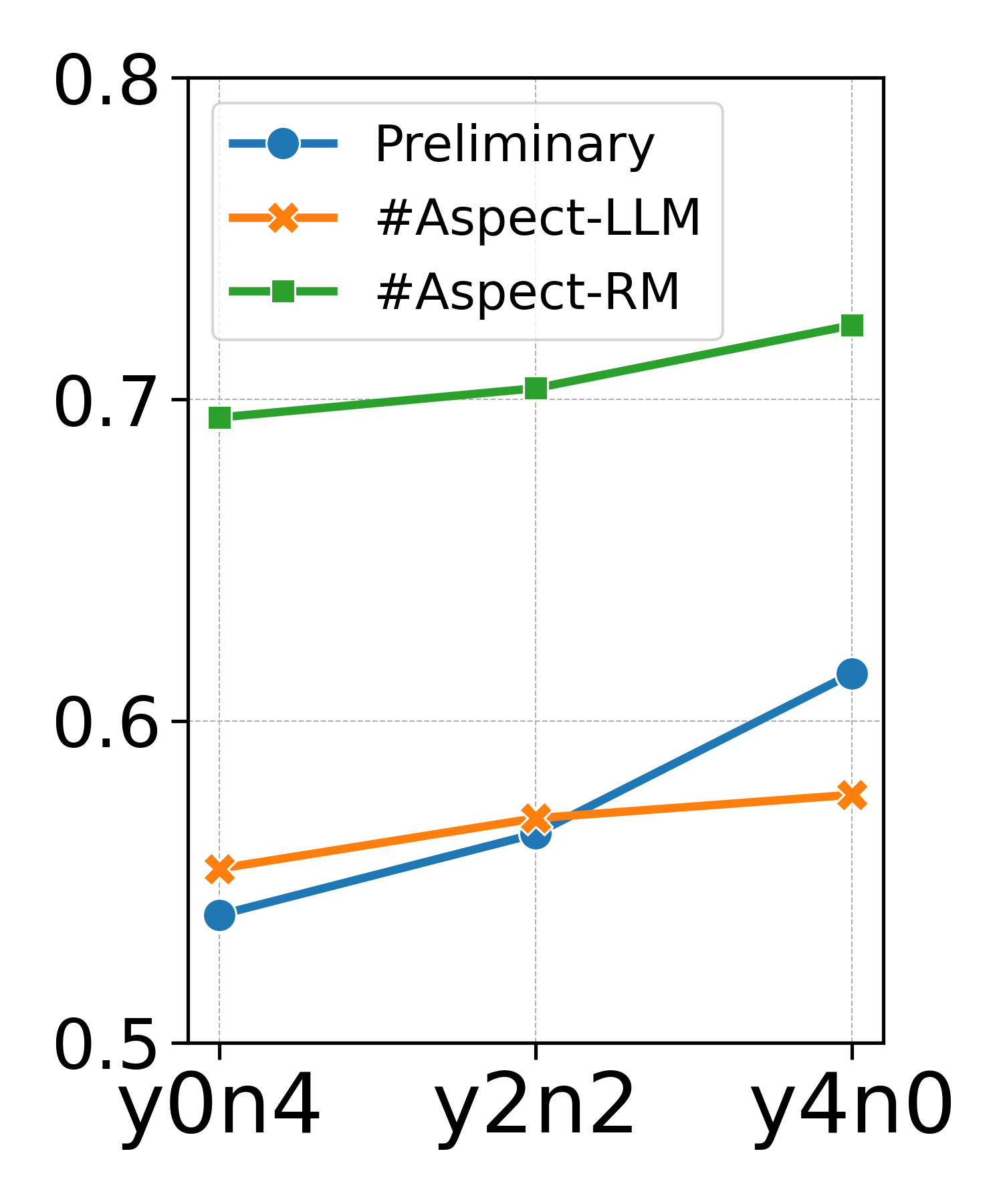}
        \caption{COVID-19-PC}
        \label{fig:sub2}
    \end{subfigure}%
    \hfill %
    \begin{subfigure}{0.157\textwidth}
        \centering
         \includegraphics[width=\linewidth]{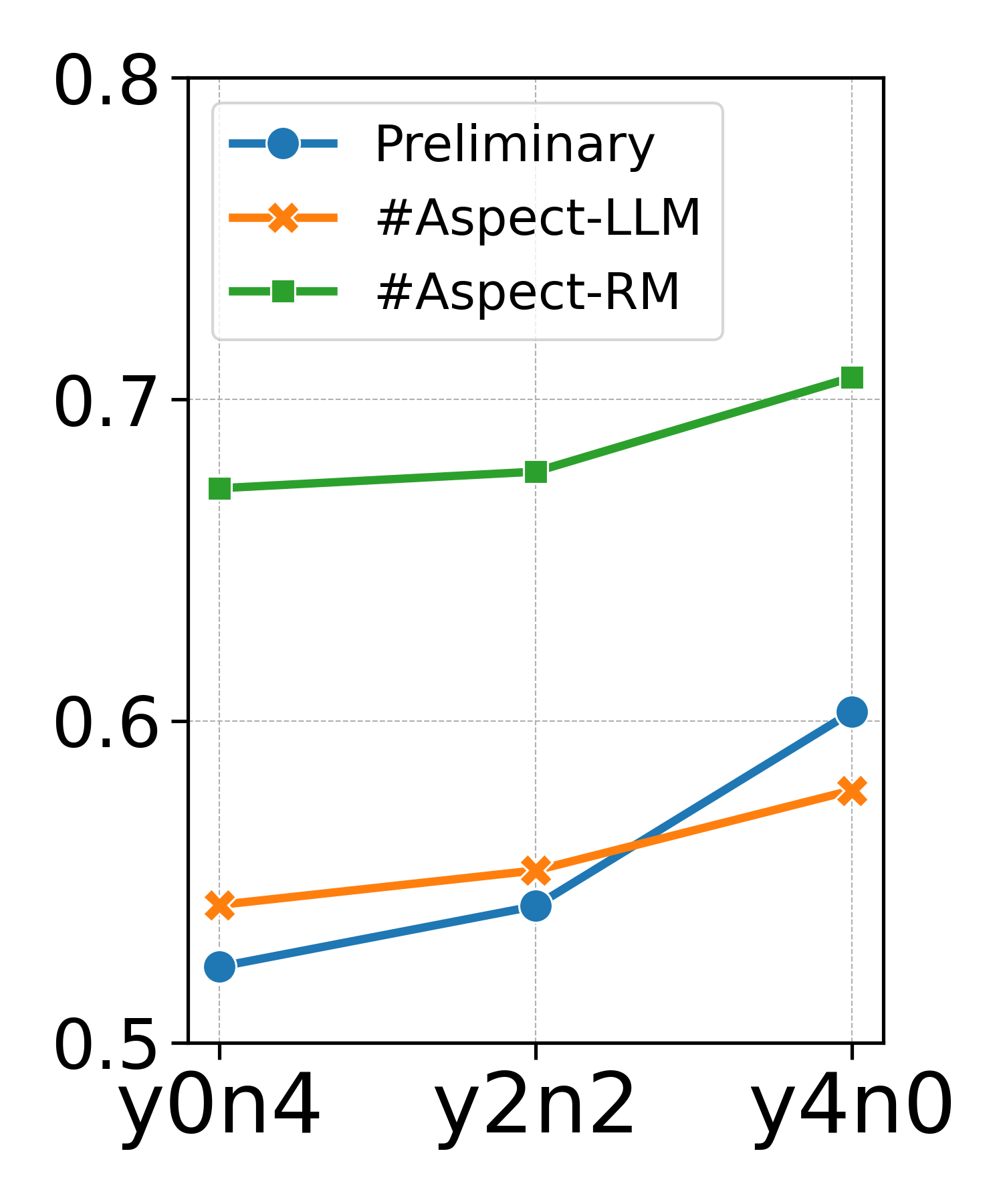}
        \caption{RW-ECT}
        \label{fig:sub3}
    \end{subfigure}%
    \caption{Average BERTScore across three provided aspect settings on the three datasets. }
     \label{fig:three_figures_combined}
\end{figure}

\begin{figure}[t]
    \centering
    \includegraphics[width=.9\linewidth]{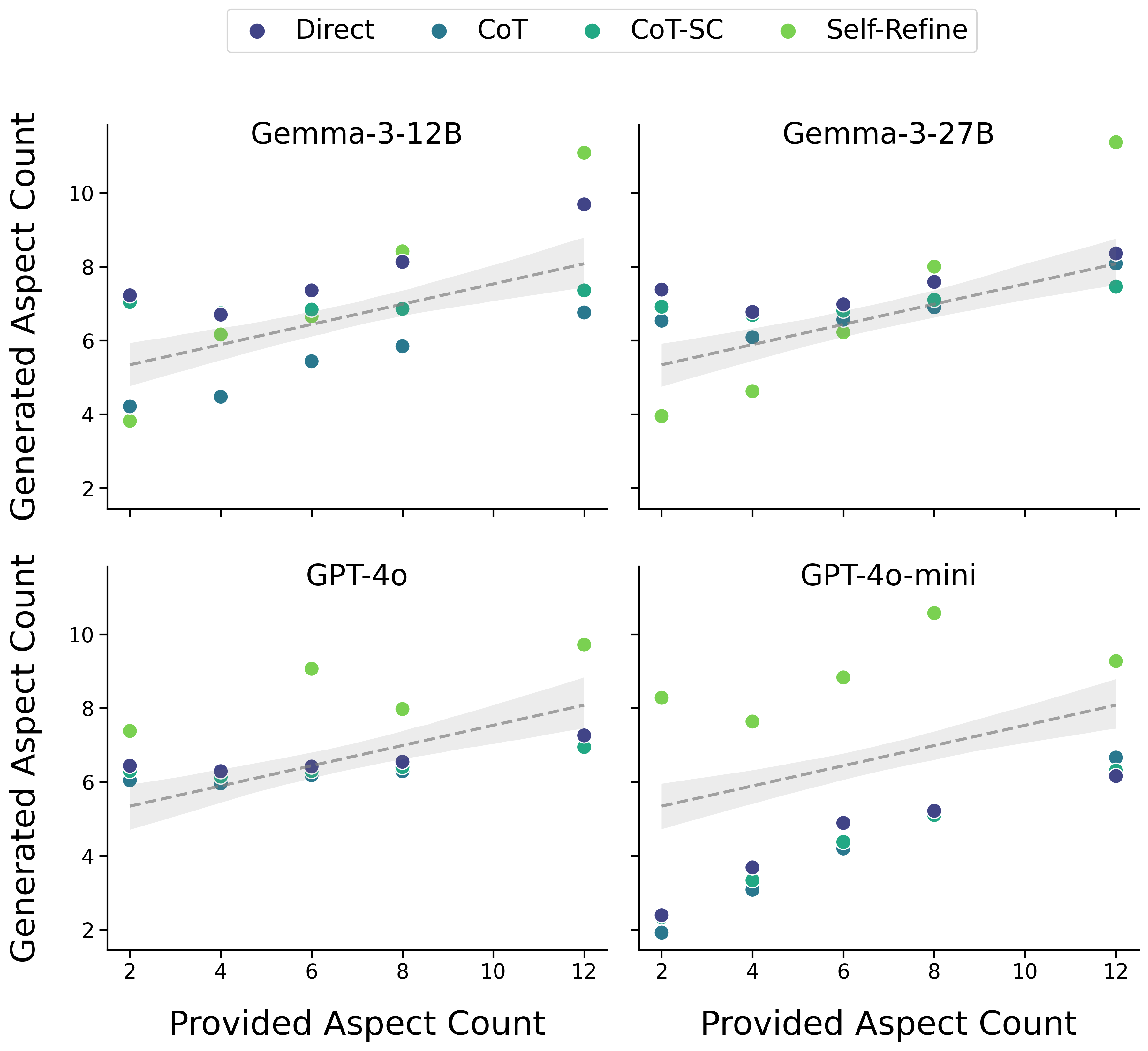}
    \caption{Relationship between the provided aspect count and the generated aspect count on ECT under the Preliminary Experiments.}
    \label{fig:ECT-avg-asp}
\end{figure}

\subsection{RQ3: Sensitivity to Provided Aspects}\label{sec:sensitivity}
We study the detailed breakdown of how the quality of initial provided aspects influences summarization performance, and conduct experiments with the three settings: completely incorrect (y0n4), a mix of correct and incorrect (y2n2), and completely correct (y4n0).
The results are reported in Figure \ref{fig:three_figures_combined}.
A clear trend emerges across all approaches: summarization quality, as measured by average BERTScore, improves as the proportion of correct aspects in the initial input increases. 
This highlights that noisy input aspects might potentially lead to inconsistencies and ultimately hinder the summarization performance of LLMs. 
In addition, Figure \ref{fig:three_figures_combined} reveals the robustness of the Regression Model approach. 
While the performance of the Preliminary Experiments is highly sensitive to the quality of the initial provided aspects, our method maintains a high BERTScore even in the most challenging scenario (y0n4). 
This demonstrates the effectiveness of \#Aspect-RM in filtering out noise and focusing on content-relevant aspects, a critical capability for real-world applications where provided aspects are often imperfect.

Furthermore, a closer examination of the total number of generated aspects, as illustrated in Figure \ref{fig:ECT-avg-asp}, unveils a positive correlation with the total number of provided aspects.
In the Preliminary Experiments, we observe that LLMs may exhibit a compliance phenomenon, where they tend to directly generate a similar number of summaries as the total number of provided aspects without sufficient critical thinking or effectively refining the aspect set.
This behavior can negatively impact summarization quality, as the model is compelled to produce content for irrelevant or non-existent aspects. Such overgeneration leads to hallucinations and factual inaccuracies, ultimately degrading the overall BERTScore, as evidenced in Figure~\ref{fig:three_figures_combined}.

\subsection{Effect of Self-Refine Thinking Steps} 
To further probe the operational dynamics of Self-Refine, we analyze its average thinking steps (Avg. Step) on each dataset.
Our findings show that both \#Aspect-LLM and \#Aspect-RM guidance encourage LLMs to take more thinking steps; however, this does not consistently lead to better performance.
This indicates that accurately estimating the number of aspects is more crucial than simply increasing the number of refinement steps.
Surprisingly, our analysis also uncovers an inverse relationship between model size and reasoning effort: smaller models consistently exhibit a higher average step than their larger counterparts, which might sometimes overthink and make unnecessary edits.
The detailed results on each dataset and LLM are provided in Appendix Table \ref{tab:combined_self_refine}.

\section{Conclusion}
This paper introduces CARPAS, a novel and practical task setting that addresses the challenge of handling imperfect input aspects in real-world ABS problems. 
To tackle this, we propose a simple yet highly effective two-stage framework that first predicts the number of relevant aspects by a regression model, and then uses this information to guide LLMs for the aspect refinement and following summarization in a content-aware manner.
Experiments on three different datasets show that our approach consistently outperforms popular prompting strategies, validating our intuition that providing explicit numerical guidance to LLMs can improve their performance in CARPAS.
We believe our results provide useful insights for similar aspect-finding applications, and multiple future research directions can be further explored, including applying our approach to areas such as structured data-to-text generation or few-shot dialogue systems, further extending the impact of our findings.
\section*{Limitations}
While our proposed framework demonstrates strong results on the newly introduced CARPAS task, several limitations remain, pointing to directions for future work.

\noindent \textbf{Dataset.}
The datasets used in our experiments are relatively small, constrained by the high cost of LLM-based data generation and the impracticality of large-scale manual annotation.
Expanding these datasets through scalable self-supervised generation or crowd-sourced pipelines could improve model robustness and generalizability.
In addition, although CARPAS was evaluated in finance and public health, the domain coverage could be further expanded, and exploration in areas such as law, education, scientific literature, or sports could better test domain-agnostic applicability and uncover domain-specific challenges in aspect refinement and summarization.

\noindent \textbf{Method.}
We note that the success of the entire framework highly depends on the accuracy of the initial aspect count prediction, which makes the subsequent ABS task brittle.
Moreover, when there are hierarchical or nested aspect structures in the documents, \#Aspect-RM may no longer be effective.
To address this, we plan to explore a more sophisticated agentic approach to identify the correct aspect counts and the aspect structure of each document in future work.

\bibliography{reference}

\begin{thebibliography}{37}
\providecommand{\natexlab}[1]{#1}

\bibitem[{Ahuja et~al.(2022)Ahuja, Xu, Gupta, Horecka, and
  Durrett}]{Aspect_news}
Ojas Ahuja, Jiacheng Xu, Akshay Gupta, Kevin Horecka, and Greg Durrett. 2022.
\newblock {ASPECTNEWS:} aspect-oriented summarization of news documents.
\newblock In \emph{{ACL} {(1)}}, pages 6494--6506. Association for
  Computational Linguistics.

\bibitem[{Amar et~al.(2023)Amar, Schiff, Ernst, Shefer, Shapira, and
  Dagan}]{OpenAsp}
Shmuel Amar, Liat Schiff, Ori Ernst, Asi Shefer, Ori Shapira, and Ido Dagan.
  2023.
\newblock Openasp: {A} benchmark for multi-document open aspect-based
  summarization.
\newblock In \emph{{EMNLP}}, pages 1967--1991. Association for Computational
  Linguistics.

\bibitem[{Angelidis and Lapata(2018)}]{summarize_opinions}
Stefanos Angelidis and Mirella Lapata. 2018.
\newblock Summarizing opinions: Aspect extraction meets sentiment prediction
  and they are both weakly supervised.
\newblock In \emph{{EMNLP}}, pages 3675--3686. Association for Computational
  Linguistics.

\bibitem[{Bi et~al.(2024)Bi, Chen, Chen, Chen, Dai, Deng, Ding, Dong, Du, Fu,
  Gao, Gao, Gao, Ge, Guan, Guo, Guo, Hao, Hao, He, Hu, Huang, Li, Li, Li, Li,
  Li, Liang, Lin, Liu, Liu, Liu, Liu, Liu, Liu, Lu, Lu, Luo, Ma, Nie, Pei,
  Piao, Qiu, Qu, Ren, Ren, Ruan, Sha, Shao, Song, Su, Sun, Sun, Tang, Wang,
  Wang, Wang, Wang, Wang, Wu, Wu, Xie, Xie, Xie, Xiong, Xu, Xu, Xu, Yang, You,
  Yu, Yu, Zhang, Zhang, Zhang, Zhang, Zhang, Zhang, Zhang, Zhang, Zhao, Zhao,
  Zhou, Zhou, Zhu, and Zou}]{deepSeek}
Xiao Bi, Deli Chen, Guanting Chen, Shanhuang Chen, Damai Dai, Chengqi Deng,
  Honghui Ding, Kai Dong, Qiushi Du, Zhe Fu, Huazuo Gao, Kaige Gao, Wenjun Gao,
  Ruiqi Ge, Kang Guan, Daya Guo, Jianzhong Guo, Guangbo Hao, Zhewen Hao, and 67
  others. 2024.
\newblock Deepseek {LLM:} scaling open-source language models with longtermism.
\newblock \emph{CoRR}, abs/2401.02954.

\bibitem[{Brown et~al.(2020)Brown, Mann, Ryder, Subbiah, Kaplan, Dhariwal,
  Neelakantan, Shyam, Sastry, Askell, Agarwal, Herbert{-}Voss, Krueger,
  Henighan, Child, Ramesh, Ziegler, Wu, Winter, Hesse, Chen, Sigler, Litwin,
  Gray, Chess, Clark, Berner, McCandlish, Radford, Sutskever, and
  Amodei}]{few-show-learner}
Tom~B. Brown, Benjamin Mann, Nick Ryder, Melanie Subbiah, Jared Kaplan,
  Prafulla Dhariwal, Arvind Neelakantan, Pranav Shyam, Girish Sastry, Amanda
  Askell, Sandhini Agarwal, Ariel Herbert{-}Voss, Gretchen Krueger, Tom
  Henighan, Rewon Child, Aditya Ramesh, Daniel~M. Ziegler, Jeffrey Wu, Clemens
  Winter, and 12 others. 2020.
\newblock Language models are few-shot learners.
\newblock In \emph{NeurIPS}.

\bibitem[{Deng et~al.(2023)Deng, Yoon, Bui, Dernoncourt, Tran, Liu, Zhao,
  Zhang, Wang, and Yu}]{meeting_abs_dataset_2023}
Zhongfen Deng, Seunghyun Yoon, Trung Bui, Franck Dernoncourt, Quan~Hung Tran,
  Shuaiqi Liu, Wenting Zhao, Tao Zhang, Yibo Wang, and Philip~S. Yu. 2023.
\newblock Aspect-based meeting transcript summarization: {A} two-stage approach
  with weak supervision on sentence classification.
\newblock In \emph{{IEEE} Big Data}, pages 636--645. {IEEE}.

\bibitem[{Dubey et~al.(2024)Dubey, Jauhri, Pandey, Kadian, Al{-}Dahle, Letman,
  Mathur, Schelten, Yang, Fan, Goyal, Hartshorn, Yang, Mitra, Sravankumar,
  Korenev, Hinsvark, Rao, Zhang, Rodriguez, Gregerson, Spataru, Rozi{\`{e}}re,
  Biron, Tang, Chern, Caucheteux, Nayak, Bi, Marra, McConnell, Keller, Touret,
  Wu, Wong, Ferrer, Nikolaidis, Allonsius, Song, Pintz, Livshits, Esiobu,
  Choudhary, Mahajan, Garcia{-}Olano, Perino, Hupkes, Lakomkin, AlBadawy,
  Lobanova, Dinan, Smith, Radenovic, Zhang, Synnaeve, Lee, Anderson, Nail,
  Mialon, Pang, Cucurell, Nguyen, Korevaar, Xu, Touvron, Zarov, Ibarra,
  Kloumann, Misra, Evtimov, Copet, Lee, Geffert, Vranes, Park, Mahadeokar,
  Shah, van~der Linde, Billock, Hong, Lee, Fu, Chi, Huang, Liu, Wang, Yu,
  Bitton, Spisak, Park, Rocca, Johnstun, Saxe, Jia, Alwala, Upasani, Plawiak,
  Li, Heafield, Stone, and et~al.}]{llama3}
Abhimanyu Dubey, Abhinav Jauhri, Abhinav Pandey, Abhishek Kadian, Ahmad
  Al{-}Dahle, Aiesha Letman, Akhil Mathur, Alan Schelten, Amy Yang, Angela Fan,
  Anirudh Goyal, Anthony Hartshorn, Aobo Yang, Archi Mitra, Archie Sravankumar,
  Artem Korenev, Arthur Hinsvark, Arun Rao, Aston Zhang, and 82 others. 2024.
\newblock The llama 3 herd of models.
\newblock \emph{CoRR}, abs/2407.21783.

\bibitem[{Feng et~al.(2025)Feng, Zhao, Li, Xiao, Wu, and Luu}]{SARESG_2025}
Yichao Feng, Shuai Zhao, Yueqiu Li, Luwei Xiao, Xiaobao Wu, and Anh~Tuan Luu.
  2025.
\newblock Aspect-based summarization with self-aspect retrieval enhanced
  generation.
\newblock \emph{CoRR}, abs/2504.13054.

\bibitem[{Frermann and Klementiev(2019)}]{MA_news}
Lea Frermann and Alexandre Klementiev. 2019.
\newblock Inducing document structure for aspect-based summarization.
\newblock In \emph{{ACL} {(1)}}, pages 6263--6273. Association for
  Computational Linguistics.

\bibitem[{Guo and Vosoughi(2024{\natexlab{a}})}]{disorder_DABS}
Xiaobo Guo and Soroush Vosoughi. 2024{\natexlab{a}}.
\newblock Disordered-dabs: {A} benchmark for dynamic aspect-based summarization
  in disordered texts.
\newblock In \emph{{EMNLP} (Findings)}, pages 416--431. Association for
  Computational Linguistics.

\bibitem[{Guo and Vosoughi(2024{\natexlab{b}})}]{MODABS}
Xiaobo Guo and Soroush Vosoughi. 2024{\natexlab{b}}.
\newblock {MODABS:} multi-objective learning for dynamic aspect-based
  summarization.
\newblock In \emph{{ACL} (Findings)}, pages 2814--2827. Association for
  Computational Linguistics.

\bibitem[{Hayashi et~al.(2021)Hayashi, Budania, Wang, Ackerson, Neervannan, and
  Neubig}]{WikiASP}
Hiroaki Hayashi, Prashant Budania, Peng Wang, Chris Ackerson, Raj Neervannan,
  and Graham Neubig. 2021.
\newblock Wikiasp: {A} dataset for multi-domain aspect-based summarization.
\newblock \emph{Trans. Assoc. Comput. Linguistics}, 9:211--225.

\bibitem[{He et~al.(2022)He, Kryscinski, McCann, Rajani, and Xiong}]{CTRLSum}
Junxian He, Wojciech Kryscinski, Bryan McCann, Nazneen Rajani, and Caiming
  Xiong. 2022.
\newblock Ctrlsum: Towards generic controllable text summarization.
\newblock In \emph{{EMNLP}}, pages 5879--5915. Association for Computational
  Linguistics.

\bibitem[{He et~al.(2023)He, Gao, and Chen}]{deberta-v3}
Pengcheng He, Jianfeng Gao, and Weizhu Chen. 2023.
\newblock Debertav3: Improving deberta using electra-style pre-training with
  gradient-disentangled embedding sharing.
\newblock In \emph{{ICLR}}. OpenReview.net.

\bibitem[{Hu et~al.(2022)Hu, Shen, Wallis, Allen{-}Zhu, Li, Wang, Wang, and
  Chen}]{lora}
Edward~J. Hu, Yelong Shen, Phillip Wallis, Zeyuan Allen{-}Zhu, Yuanzhi Li,
  Shean Wang, Lu~Wang, and Weizhu Chen. 2022.
\newblock Lora: Low-rank adaptation of large language models.
\newblock In \emph{{ICLR}}. OpenReview.net.

\bibitem[{Hurst et~al.(2024)Hurst, Lerer, Goucher, Perelman, Ramesh, Clark,
  Ostrow, Welihinda, Hayes, Radford, Madry, Baker{-}Whitcomb, Beutel, Borzunov,
  Carney, Chow, Kirillov, Nichol, Paino, Renzin, Passos, Kirillov, Christakis,
  Conneau, Kamali, Jabri, Moyer, Tam, Crookes, Tootoonchian, Kumar, Vallone,
  Karpathy, Braunstein, Cann, Codispoti, Galu, Kondrich, Tulloch, Mishchenko,
  Baek, Jiang, Pelisse, Woodford, Gosalia, Dhar, Pantuliano, Nayak, Oliver,
  Zoph, Ghorbani, Leimberger, Rossen, Sokolowsky, Wang, Zweig, Hoover, Samic,
  McGrew, Spero, Giertler, Cheng, Lightcap, Walkin, Quinn, Guarraci, Hsu,
  Kellogg, Eastman, Lugaresi, Wainwright, Bassin, Hudson, Chu, Nelson, Li,
  Shern, Conger, Barette, Voss, Ding, Lu, Zhang, Beaumont, Hallacy, Koch,
  Gibson, Kim, Choi, McLeavey, Hesse, Fischer, Winter, Czarnecki, Jarvis, Wei,
  Koumouzelis, and Sherburn}]{gpt_4o}
Aaron Hurst, Adam Lerer, Adam~P. Goucher, Adam Perelman, Aditya Ramesh, Aidan
  Clark, AJ~Ostrow, Akila Welihinda, Alan Hayes, Alec Radford, Aleksander
  Madry, Alex Baker{-}Whitcomb, Alex Beutel, Alex Borzunov, Alex Carney, Alex
  Chow, Alex Kirillov, Alex Nichol, Alex Paino, and 79 others. 2024.
\newblock Gpt-4o system card.
\newblock \emph{CoRR}, abs/2410.21276.

\bibitem[{Kamath et~al.(2025)Kamath, Ferret, Pathak, Vieillard, Merhej, Perrin,
  Matejovicova, Ram{\'{e}}, Rivi{\`{e}}re, Rouillard, Mesnard, Cideron, Grill,
  Ramos, Yvinec, Casbon, Pot, Penchev, Liu, Visin, Kenealy, Beyer, Zhai,
  Tsitsulin, Busa{-}Fekete, Feng, Sachdeva, Coleman, Gao, Mustafa, Barr,
  Parisotto, Tian, Eyal, Cherry, Peter, Sinopalnikov, Bhupatiraju, Agarwal,
  Kazemi, Malkin, Kumar, Vilar, Brusilovsky, Luo, Steiner, Friesen, Sharma,
  Sharma, Gilady, Goedeckemeyer, Saade, Kolesnikov, Bendebury, Abdagic, Vadi,
  Gy{\"{o}}rgy, Pinto, Das, Bapna, Miech, Yang, Paterson, Shenoy, Chakrabarti,
  Piot, Wu, Shahriari, Petrini, Chen, Lan, Choquette{-}Choo, Carey, Brick,
  Deutsch, Eisenbud, Cattle, Cheng, Paparas, Sreepathihalli, Reid, Tran, Zelle,
  Noland, Huizenga, Kharitonov, Liu, Amirkhanyan, Cameron, Hashemi,
  Klimczak{-}Plucinska, Singh, Mehta, Lehri, Hazimeh, Ballantyne, Szpektor, and
  Nardini}]{gemma_3}
Aishwarya Kamath, Johan Ferret, Shreya Pathak, Nino Vieillard, Ramona Merhej,
  Sarah Perrin, Tatiana Matejovicova, Alexandre Ram{\'{e}}, Morgane
  Rivi{\`{e}}re, Louis Rouillard, Thomas Mesnard, Geoffrey Cideron,
  Jean{-}Bastien Grill, Sabela Ramos, Edouard Yvinec, Michelle Casbon, Etienne
  Pot, Ivo Penchev, Ga{\"{e}}l Liu, and 79 others. 2025.
\newblock Gemma 3 technical report.
\newblock \emph{CoRR}, abs/2503.19786.

\bibitem[{Kuhn(1955)}]{kuhn1955hungarian}
Harold~W Kuhn. 1955.
\newblock The hungarian method for the assignment problem.
\newblock \emph{Naval research logistics quarterly}, 2(1-2):83--97.

\bibitem[{Kunneman et~al.(2018)Kunneman, Wubben, van~den Bosch, and
  Krahmer}]{Summary_product}
Florian Kunneman, Sander Wubben, Antal van~den Bosch, and Emiel Krahmer. 2018.
\newblock Aspect-based summarization of pros and cons in unstructured product
  reviews.
\newblock In \emph{{COLING}}, pages 2219--2229. Association for Computational
  Linguistics.

\bibitem[{Li et~al.(2023)Li, Chowdhury, and Chaturvedi}]{TokenCluster}
Haoyuan Li, Somnath Basu~Roy Chowdhury, and Snigdha Chaturvedi. 2023.
\newblock Aspect-aware unsupervised extractive opinion summarization.
\newblock In \emph{{ACL} (Findings)}, pages 12662--12678. Association for
  Computational Linguistics.

\bibitem[{Lin(2004)}]{rouge}
Chin-Yew Lin. 2004.
\newblock \href {https://aclanthology.org/W04-1013/} {{ROUGE}: A package for
  automatic evaluation of summaries}.
\newblock In \emph{Text Summarization Branches Out}, pages 74--81, Barcelona,
  Spain. Association for Computational Linguistics.

\bibitem[{Loshchilov and Hutter(2019)}]{adamw}
Ilya Loshchilov and Frank Hutter. 2019.
\newblock \href {https://openreview.net/forum?id=Bkg6RiCqY7} {Decoupled weight
  decay regularization}.
\newblock In \emph{International Conference on Learning Representations}.

\bibitem[{Madaan et~al.(2023)Madaan, Tandon, Gupta, Hallinan, Gao, Wiegreffe,
  Alon, Dziri, Prabhumoye, Yang, Gupta, Majumder, Hermann, Welleck,
  Yazdanbakhsh, and Clark}]{self_refine}
Aman Madaan, Niket Tandon, Prakhar Gupta, Skyler Hallinan, Luyu Gao, Sarah
  Wiegreffe, Uri Alon, Nouha Dziri, Shrimai Prabhumoye, Yiming Yang, Shashank
  Gupta, Bodhisattwa~Prasad Majumder, Katherine Hermann, Sean Welleck, Amir
  Yazdanbakhsh, and Peter Clark. 2023.
\newblock Self-refine: Iterative refinement with self-feedback.
\newblock In \emph{NeurIPS}.

\bibitem[{Maddela et~al.(2022)Maddela, Kulkarni, and
  Preotiuc{-}Pietro}]{EntSum}
Mounica Maddela, Mayank Kulkarni, and Daniel Preotiuc{-}Pietro. 2022.
\newblock Entsum: {A} data set for entity-centric extractive summarization.
\newblock In \emph{{ACL} {(1)}}, pages 3355--3366. Association for
  Computational Linguistics.

\bibitem[{Meng et~al.(2021)Meng, Thaker, Zhang, Dong, Yuan, Wang, and
  He}]{FacetedSum}
Rui Meng, Khushboo Thaker, Lei Zhang, Yue Dong, Xingdi Yuan, Tong Wang, and
  Daqing He. 2021.
\newblock Bringing structure into summaries: a faceted summarization dataset
  for long scientific documents.
\newblock In \emph{{ACL/IJCNLP} {(2)}}, pages 1080--1089. Association for
  Computational Linguistics.

\bibitem[{Minaee et~al.(2024)Minaee, Mikolov, Nikzad, Chenaghlu, Socher,
  Amatriain, and Gao}]{llm_survey}
Shervin Minaee, Tom{\'{a}}s Mikolov, Narjes Nikzad, Meysam Chenaghlu, Richard
  Socher, Xavier Amatriain, and Jianfeng Gao. 2024.
\newblock Large language models: {A} survey.
\newblock \emph{CoRR}, abs/2402.06196.

\bibitem[{Mullick et~al.(2024)Mullick, Bose, Saha, Bhowmick, Vempaty, Goyal,
  Ganguly, Dey, and Kokku}]{llm_abs}
Ankan Mullick, Sombit Bose, Rounak Saha, Ayan~Kumar Bhowmick, Aditya Vempaty,
  Pawan Goyal, Niloy Ganguly, Prasenjit Dey, and Ravi Kokku. 2024.
\newblock Leveraging the power of llms: {A} fine-tuning approach for
  high-quality aspect-based summarization.
\newblock \emph{CoRR}, abs/2408.02584.

\bibitem[{Takeshita et~al.(2024)Takeshita, Green, Reinig, Eckert, and
  Ponzetto}]{ACLSum}
Sotaro Takeshita, Tommaso Green, Ines Reinig, Kai Eckert, and Simone~Paolo
  Ponzetto. 2024.
\newblock Aclsum: {A} new dataset for aspect-based summarization of scientific
  publications.
\newblock In \emph{{NAACL-HLT}}, pages 6660--6675. Association for
  Computational Linguistics.

\bibitem[{Tan et~al.(2020)Tan, Qin, Xing, and Hu}]{AnyAspect}
Bowen Tan, Lianhui Qin, Eric~P. Xing, and Zhiting Hu. 2020.
\newblock Summarizing text on any aspects: {A} knowledge-informed
  weakly-supervised approach.
\newblock In \emph{{EMNLP} {(1)}}, pages 6301--6309. Association for
  Computational Linguistics.

\bibitem[{Tian et~al.(2025)Tian, Tang, Wang, Yen, and
  Peng}]{Template-based_Financial}
Yong{-}En Tian, Yu{-}Chien Tang, Kuang{-}Da Wang, An{-}Zi Yen, and Wen{-}Chih
  Peng. 2025.
\newblock Template-based financial report generation in agentic and decomposed
  information retrieval.
\newblock In \emph{{SIGIR}}, pages 2706--2710. {ACM}.

\bibitem[{Wang et~al.(2023)Wang, Wei, Schuurmans, Le, Chi, Narang, Chowdhery,
  and Zhou}]{CoT_SC}
Xuezhi Wang, Jason Wei, Dale Schuurmans, Quoc~V. Le, Ed~H. Chi, Sharan Narang,
  Aakanksha Chowdhery, and Denny Zhou. 2023.
\newblock Self-consistency improves chain of thought reasoning in language
  models.
\newblock In \emph{{ICLR}}. OpenReview.net.

\bibitem[{Wei et~al.(2022)Wei, Wang, Schuurmans, Bosma, Ichter, Xia, Chi, Le,
  and Zhou}]{CoT}
Jason Wei, Xuezhi Wang, Dale Schuurmans, Maarten Bosma, Brian Ichter, Fei Xia,
  Ed~H. Chi, Quoc~V. Le, and Denny Zhou. 2022.
\newblock Chain-of-thought prompting elicits reasoning in large language
  models.
\newblock In \emph{NeurIPS}.

\bibitem[{Yang et~al.(2023{\natexlab{a}})Yang, Li, Zhang, Chen, and
  Cheng}]{chatGPT_QFS_ABS}
Xianjun Yang, Yan Li, Xinlu Zhang, Haifeng Chen, and Wei Cheng.
  2023{\natexlab{a}}.
\newblock Exploring the limits of chatgpt for query or aspect-based text
  summarization.
\newblock \emph{CoRR}, abs/2302.08081.

\bibitem[{Yang et~al.(2023{\natexlab{b}})Yang, Song, Cho, Wang, Pan, Petzold,
  and Yu}]{OASum}
Xianjun Yang, Kaiqiang Song, Sangwoo Cho, Xiaoyang Wang, Xiaoman Pan, Linda~R.
  Petzold, and Dong Yu. 2023{\natexlab{b}}.
\newblock Oasum: Large-scale open domain aspect-based summarization.
\newblock In \emph{{ACL} (Findings)}, pages 4381--4401. Association for
  Computational Linguistics.

\bibitem[{Zhang et~al.(2020)Zhang, Kishore, Wu, Weinberger, and
  Artzi}]{BERTScore}
Tianyi Zhang, Varsha Kishore, Felix Wu, Kilian~Q. Weinberger, and Yoav Artzi.
  2020.
\newblock Bertscore: Evaluating text generation with {BERT}.
\newblock In \emph{{ICLR}}. OpenReview.net.

\bibitem[{Zhang et~al.(2025)Zhang, Li, Long, Zhang, Lin, Yang, Xie, Yang, Liu,
  Lin, Huang, and Zhou}]{qwe3_embedding}
Yanzhao Zhang, Mingxin Li, Dingkun Long, Xin Zhang, Huan Lin, Baosong Yang,
  Pengjun Xie, An~Yang, Dayiheng Liu, Junyang Lin, Fei Huang, and Jingren Zhou.
  2025.
\newblock Qwen3 embedding: Advancing text embedding and reranking through
  foundation models.
\newblock \emph{CoRR}, abs/2506.05176.

\bibitem[{Zhu et~al.(2021)Zhu, Tu, Shi, Li, Hou, and Cui}]{TWAG}
Fangwei Zhu, Shangqing Tu, Jiaxin Shi, Juanzi Li, Lei Hou, and Tong Cui. 2021.
\newblock {TWAG:} {A} topic-guided wikipedia abstract generator.
\newblock In \emph{{ACL/IJCNLP} {(1)}}, pages 4623--4635. Association for
  Computational Linguistics.

\end{thebibliography}

\appendix

\section{Full Results}
\label{app:full_results}
\subsection{Computing Infrastructure}
Our experimental setup consisted of a machine running Ubuntu 20.04.6 LTS (Linux kernel 5.15) with an AMD Ryzen Threadripper 3960X CPU, 251 GB of RAM, and two NVIDIA GeForce RTX 3090 GPUs. Key software versions include Python 3.12.2, CUDA 12.2, and PyTorch 2.2.0.

\subsection{Quantitative Results}
We leave the prompt templates in our code.\footnote{\url{https://anonymous.4open.science/r/CARPAS-537E/README.md}}
Table \ref{tab:ect-all}, \ref{tab:covid_all}, and \ref{tab:rw-all} present the complete evaluation metrics, including BERTScore, ROUGE-1, ROUGE-2, and ROUGE-L, for the three datasets.
As previously discussed, the \#Aspect-RM outperforms both the Preliminary Experiments and the \#Aspect-LLM. 
Furthermore, GPT-based models achieve their best performance when paired with the Self-Refine strategy.

\begin{figure}[t]
    \centering
    \includegraphics[width=\linewidth]{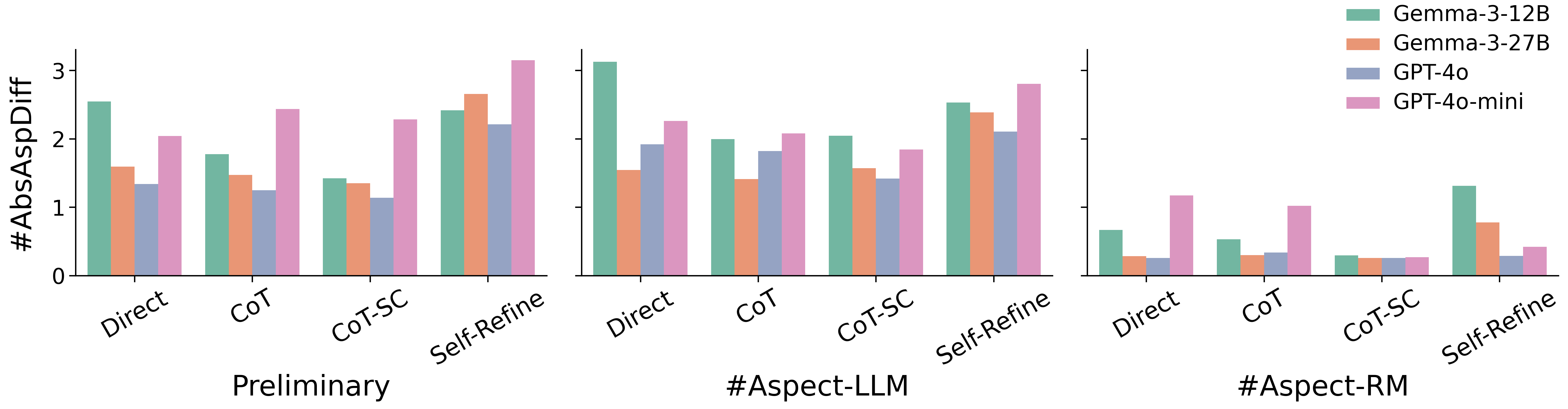}
    \caption{Aspect count error (\#AbsAspDiff) across different methods and prompting strategies on ECT.}
    \label{fig:ECT-asp}
\end{figure}

\begin{figure}[t]
    \centering
    \includegraphics[width=\linewidth]{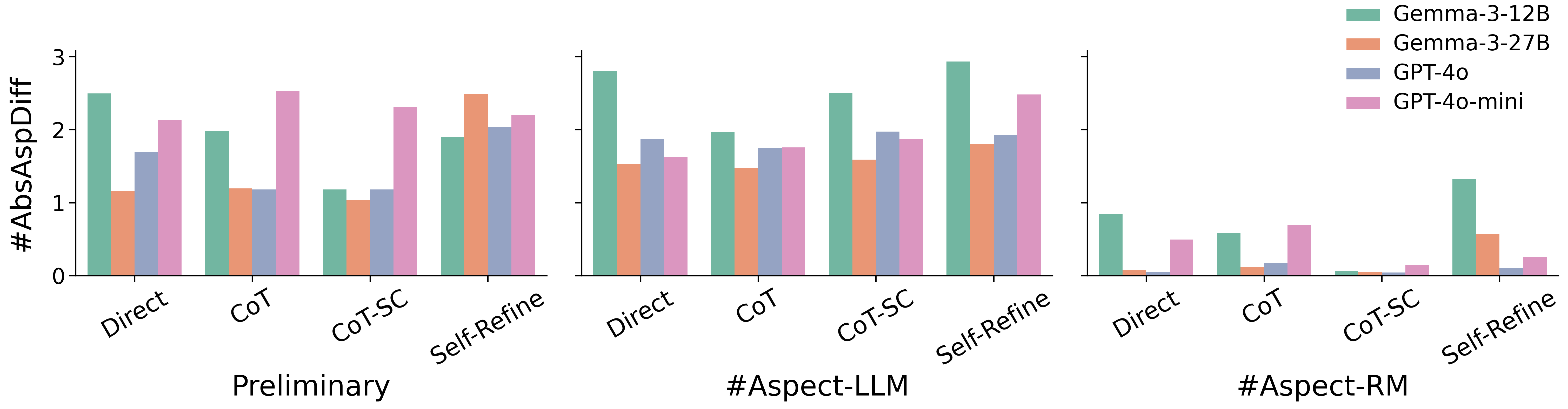}
    \caption{Aspect count error (\#AbsAspDiff) across different methods and prompting strategies on COVID-19-PC.}
    \label{fig:COVID-asp}
\end{figure}

\begin{figure}[t]
    \centering
    \includegraphics[width=\linewidth]{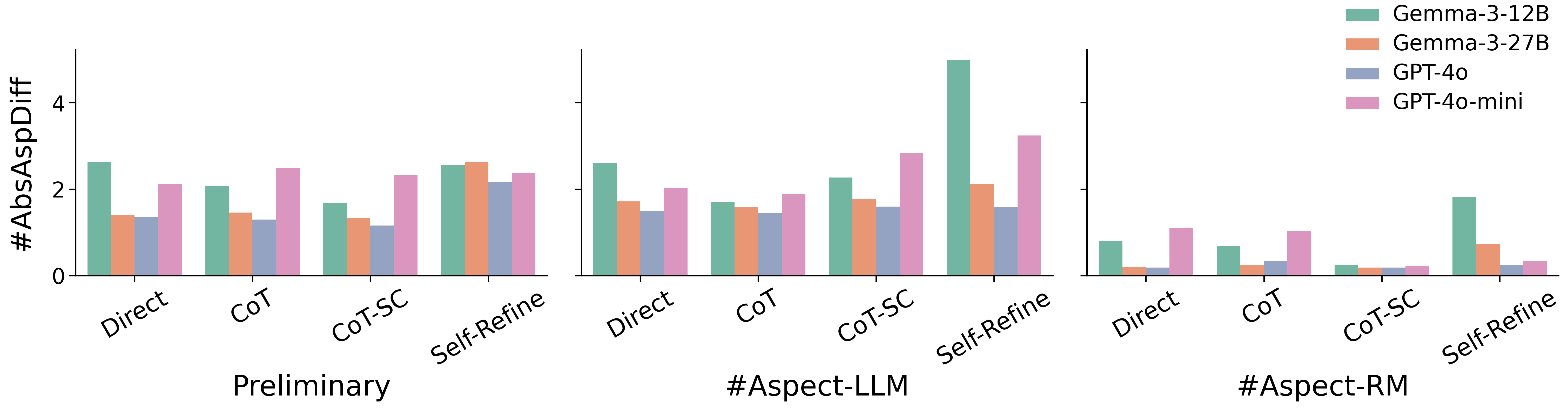}
    \caption{Aspect count error (\#AbsAspDiff) across different methods and prompting strategies on RW-ECT.}
    \label{fig:RW-asp}
\end{figure}

The primary driver of this performance gain is the model's enhanced ability to identify the correct number of aspects.
As illustrated in Figure \ref{fig:ECT-asp}, \#Aspect-RM achieves the lowest \#AbsAspDiff by a large margin, reducing by an average of 71.7\% in ECT and 82.4\% in COVID-19-PC compared to the Preliminary Experiments. 
In contrast, simply prompting the LLMs for an aspect number, as done in \#Aspect-LLM, fails to yield improvements over the Preliminary Experiments baseline due to inaccurate aspect number estimation. 
Figure \ref{fig:ECT-asp} also highlights the severity of this issue in baseline methods: in both the Preliminary Experiments and \#Aspect-LLM results, the generated aspect count deviates from the ground truth by at least one aspect on average, and sometimes by as many as two or three. 
This consistent overestimation leads to the generation of superfluous aspects and summaries, underscoring the benefit of explicit aspect number guidance to LLMs in avoiding both the omission of relevant aspects and the hallucination of irrelevant ones.
Similar trends are consistently observed on COVID-19-PC and RW-ECT, as shown in Figures~\ref{fig:COVID-asp} and \ref{fig:RW-asp}, demonstrating that the advantage of \#Aspect-RM in reducing aspect count error generalizes across datasets.
Consequently, this precision leads to consistently high-quality summaries, with its strong BERTScore performance holding true across all provided aspect settings (Figure \ref{bert-setting}).

\begin{table*}[ht]
\centering
\small
\setlength{\tabcolsep}{1mm}
\resizebox{\textwidth}{!}{%
\begin{tabular}{c c
                c c c c
                c c c c
                c c c c
                c c c c}
\toprule
\multirow{2}{*}{} & \multirow{2}{*}{\textbf{Prompt}} 
& \multicolumn{4}{c}{\textbf{Gemma-3-12B}} 
& \multicolumn{4}{c}{\textbf{Gemma-3-27B}}
& \multicolumn{4}{c}{\textbf{GPT-4o-mini}}
& \multicolumn{4}{c}{\textbf{GPT-4o}} \\
\cmidrule(lr){3-6} \cmidrule(lr){7-10} \cmidrule(lr){11-14} \cmidrule(lr){15-18}
& & \textbf{BERT} & \textbf{R-1} & \textbf{R-2} & \textbf{R-L} & \textbf{BERT} & \textbf{R-1} & \textbf{R-2} & \textbf{R-L} & \textbf{BERT} & \textbf{R-1} & \textbf{R-2} & \textbf{R-L} & \textbf{BERT} & \textbf{R-1} & \textbf{R-2} & \textbf{R-L} \\
\midrule
\multirow{4}{*}{Preliminary} & Direct & 56.1 & 39.9 & 26.2 & 32.6 & 62.6 & 42.4 & 25.3 & 33.6 & 50.9 & 31.3 & 16.3 & 23.8 & 62.7 & 40.7 & 22.7 & 31.7 \\
 & CoT & 58.1 & 39.8 & 25.9 & 32.8 & 63.5 & 42.1 & 25.7 & 34.3 & 45.3 & 26.6 & 14.5 & 21.0 & 61.6 & 35.9 & 18.9 & 28.3 \\
 & CoT-SC & 63.6 & 43.3 & 26.9 & 34.1 & 63.1 & 39.8 & 22.1 & 31.0 & 48.5 & 29.8 & 16.7 & 23.4 & 63.7 & 39.2 & 21.4 & 30.6 \\
 & Self-Refine & 56.6 & 39.7 & 25.3 & 32.2 & 51.7 & 35.2 & 21.4 & 28.3 & 53.2 & 33.2 & 18.0 & 25.7 & 56.6 & 35.5 & 18.1 & 26.6 \\
\midrule
\multirow{4}{*}{\#Aspect-LLM} & Direct & 53.5 & 37.7 & 24.5 & 30.7 & 61.6 & 40.7 & 23.5 & 32.0 & 50.9 & 30.5 & 15.5 & 23.0 & 56.3 & 35.9 & 19.5 & 27.6 \\
 & CoT & 59.5 & 40.1 & 25.4 & 32.8 & 62.8 & 40.4 & 23.8 & 32.6 & 52.3 & 30.0 & 16.0 & 23.5 & 55.8 & 31.7 & 16.4 & 25.1 \\
 & CoT-SC & 59.2 & 40.0 & 24.4 & 31.5 & 60.7 & 37.9 & 20.9 & 29.6 & 56.1 & 33.4 & 18.1 & 26.0 & 60.8 & 37.1 & 19.7 & 28.6 \\
 & Self-Refine & 56.1 & 39.2 & 25.2 & 32.0 & 53.2 & 36.3 & 21.9 & 29.0 & 51.7 & 33.8 & 19.3 & 26.6 & 55.3 & 37.3 & 20.9 & 28.7 \\
\midrule
\multirow{4}{*}{\#Aspect-RM} & Direct & 70.7 & \textbf{48.4} & \textbf{30.4} & \textbf{38.6} & \underline{72.9} & \textbf{47.5} & 26.6 & 36.7 & 59.7 & 35.5 & 17.6 & 26.7 & \underline{71.5} & \underline{44.3} & \underline{23.6} & \underline{34.0} \\
 & CoT & \underline{71.2} & \underline{47.2} & \underline{29.3} & \underline{38.3} & \textbf{73.5} & 47.1 & \textbf{27.3} & \textbf{37.7} & 61.4 & 34.8 & 18.3 & 27.3 & 69.3 & 38.6 & 19.8 & 30.3 \\
 & CoT-SC & \textbf{72.0} & 46.8 & 26.8 & 35.7 & 71.9 & 44.1 & 23.1 & 33.6 & \underline{69.9} & \underline{40.5} & \underline{21.1} & \underline{31.2} & 71.0 & 42.1 & 21.8 & 32.3 \\
 & Self-Refine & 65.2 & 43.1 & 25.7 & 34.2 & 70.6 & \underline{47.2} & \underline{26.8} & \underline{37.0} & \textbf{70.6} & \textbf{44.0} & \textbf{23.4} & \textbf{33.5} & \textbf{71.9} & \textbf{45.9} & \textbf{23.9} & \textbf{34.4} \\
\bottomrule

\end{tabular}
}
\caption{Results (×100\%) on ECT, with BERT standing for BERTScore, R-1 standing for ROUGE-1, R-2 standing for ROUGE-2, and R-L standing for ROUGE-L.
The highest results for each metric are in boldface, while the second-best are underlined.}
\label{tab:ect-all}
\end{table*}

\begin{table*}[ht]
\centering
\setlength{\tabcolsep}{1mm}
\small
\resizebox{\textwidth}{!}{%
\begin{tabular}{c c
                c c c c
                c c c c
                c c c c
                c c c c}
\toprule
\multirow{2}{*}{} & \multirow{2}{*}{\textbf{Prompt}} 
& \multicolumn{4}{c}{\textbf{Gemma-3-12B}} 
& \multicolumn{4}{c}{\textbf{Gemma-3-27B}}
& \multicolumn{4}{c}{\textbf{GPT-4o-mini}}
& \multicolumn{4}{c}{\textbf{GPT-4o}} \\
\cmidrule(lr){3-6} \cmidrule(lr){7-10} \cmidrule(lr){11-14} \cmidrule(lr){15-18}
& & \textbf{BERT} & \textbf{R-1} & \textbf{R-2} & \textbf{R-L} & \textbf{BERT} & \textbf{R-1} & \textbf{R-2} & \textbf{R-L} & \textbf{BERT} & \textbf{R-1} & \textbf{R-2} & \textbf{R-L} & \textbf{BERT} & \textbf{R-1} & \textbf{R-2} & \textbf{R-L} \\
\midrule
\multirow{4}{*}{Preliminary} & Direct & 56.9 & 41.2 & 26.4 & 33.8 & 64.6 & 44.4 & 25.7 & 35.4 & 47.9 & 28.2 & 13.0 & 20.9 & 57.6 & 35.7 & 18.4 & 27.8 \\
 & CoT & 55.2 & 38.5 & 24.2 & 31.8 & 64.2 & 43.2 & 25.5 & 35.3 & 40.9 & 22.8 & 11.0 & 17.6 & 60.0 & 34.1 & 17.0 & 27.0 \\
 & CoT-SC & 66.2 & 47.5 & 29.7 & 38.1 & 65.2 & 43.4 & 24.5 & 34.7 & 46.2 & 27.8 & 14.2 & 21.4 & 62.2 & 38.6 & 20.4 & 30.4 \\
 & Self-Refine & 60.2 & 43.0 & 26.9 & 35.2 & 52.2 & 36.6 & 21.8 & 29.6 & 52.6 & 34.2 & 18.3 & 26.7 & 56.0 & 36.9 & 19.4 & 28.9 \\
\midrule
\multirow{4}{*}{\#Aspect-LLM} & Direct & 54.2 & 39.0 & 24.8 & 32.2 & 61.0 & 41.6 & 23.9 & 33.2 & 55.1 & 32.1 & 14.4 & 23.6 & 56.2 & 34.3 & 17.1 & 26.5 \\
 & CoT & 58.8 & 41.0 & 25.8 & 34.0 & 61.4 & 41.3 & 24.3 & 33.6 & 53.4 & 29.1 & 13.7 & 22.4 & 55.8 & 31.1 & 15.1 & 24.5 \\
 & CoT-SC & 55.6 & 39.3 & 24.2 & 31.7 & 59.9 & 39.3 & 21.5 & 31.2 & 55.3 & 31.9 & 15.8 & 24.5 & 56.1 & 34.1 & 17.6 & 26.8 \\
 & Self-Refine & 53.9 & 37.8 & 23.3 & 30.8 & 60.4 & 42.3 & 24.9 & 34.3 & 54.0 & 33.9 & 17.6 & 26.3 & 57.0 & 36.2 & 18.3 & 28.1 \\
\midrule
\multirow{4}{*}{\#Aspect-RM} & Direct & 69.7 & \underline{49.1} & \underline{30.1} & \underline{39.8} & \textbf{74.4} & \underline{49.4} & 27.5 & 38.9 & 64.2 & 36.4 & 15.6 & 26.5 & 71.0 & 42.4 & 20.4 & 32.4 \\
 & CoT & \underline{70.1} & 47.9 & 29.5 & 39.4 & \underline{74.2} & 48.9 & \underline{27.9} & \underline{39.4} & 61.6 & 33.3 & 15.3 & 25.4 & 69.0 & 38.2 & 18.4 & 30.1 \\
 & CoT-SC & \textbf{75.9} & \textbf{52.3} & \textbf{30.9} & \textbf{41.1} & 74.0 & 47.6 & 25.2 & 37.2 & \underline{68.7} & \underline{38.8} & \underline{18.5} & \underline{29.5} & \underline{72.4} & \underline{43.7} & \underline{22.0} & \underline{33.7} \\
 & Self-Refine & 66.9 & 46.8 & 28.5 & 38.1 & 72.4 & \textbf{50.5} & \textbf{29.1} & \textbf{40.3} & \textbf{71.2} & \textbf{44.3} & \textbf{22.4} & \textbf{34.1} & \textbf{73.1} & \textbf{46.6} & \textbf{23.3} & \textbf{35.9} \\
\bottomrule
\end{tabular}
}
\caption{Results (×100\%) on COVID-19-PC.}
\label{tab:covid_all}
\end{table*}

\begin{table*}[ht]
\centering
\setlength{\tabcolsep}{1mm}
\small
\resizebox{\textwidth}{!}{%
\begin{tabular}{c c
                c c c c
                c c c c
                c c c c
                c c c c}
\toprule
\multirow{2}{*}{} & \multirow{2}{*}{Prompt} 
& \multicolumn{4}{c}{\textbf{Gemma-3-12B}} 
& \multicolumn{4}{c}{\textbf{Gemma-3-27B}}
& \multicolumn{4}{c}{\textbf{GPT-4o-mini}}
& \multicolumn{4}{c}{\textbf{GPT-4o}} \\
\cmidrule(lr){3-6} \cmidrule(lr){7-10} \cmidrule(lr){11-14} \cmidrule(lr){15-18}
& & \textbf{BERT} & \textbf{R-1} & \textbf{R-2} & \textbf{R-L} & \textbf{BERT} & \textbf{R-1} & \textbf{R-2} & \textbf{R-L} & \textbf{BERT} & \textbf{R-1} & \textbf{R-2} & \textbf{R-L} & \textbf{BERT} & \textbf{R-1} & \textbf{R-2} & \textbf{R-L} \\
\midrule
\multirow{4}{*}{Preliminary} & Direct & 54.9 & 37.0 & 22.8 & 29.1 & 62.0 & 39.3 & 21.8 & 30.1 & 48.3 & 27.2 & 12.6 & 19.8 & 61.0 & 36.9 & 19.7 & 27.9 \\
 & CoT & 54.6 & 36.4 & 22.7 & 29.5 & 62.3 & 39.5 & 23.2 & 31.6 & 43.0 & 24.1 & 12.2 & 18.3 & 59.8 & 33.2 & 17.0 & 25.6 \\
 & CoT-SC & 61.2 & 39.9 & 23.8 & 30.6 & 62.1 & 37.9 & 20.6 & 29.0 & 46.8 & 26.9 & 13.7 & 20.2 & 62.5 & 37.0 & 19.4 & 28.2 \\
 & Self-Refine & 55.4 & 37.1 & 22.6 & 29.2 & 51.4 & 34.1 & 20.0 & 26.6 & 51.4 & 32.2 & 16.9 & 24.0 & 54.7 & 34.7 & 18.4 & 25.9 \\
\midrule
\multirow{4}{*}{\#Aspect-LLM} & Direct & 56.0 & 36.9 & 22.3 & 29.1 & 59.0 & 36.8 & 20.0 & 27.9 & 51.1 & 28.6 & 13.3 & 20.7 & 58.7 & 34.7 & 17.6 & 25.7 \\
 & CoT & 60.5 & 39.0 & 23.7 & 31.1 & 60.4 & 37.4 & 21.4 & 29.7 & 52.9 & 28.6 & 14.0 & 21.6 & 58.2 & 31.3 & 15.5 & 24.1 \\
 & CoT-SC & 56.3 & 35.1 & 19.7 & 26.6 & 57.7 & 34.0 & 17.8 & 25.7 & 51.6 & 28.6 & 14.2 & 21.6 & 58.3 & 33.6 & 17.0 & 25.3 \\
 & Self-Refine & 45.4 & 28.2 & 16.1 & 21.9 & 58.4 & 37.9 & 21.6 & 29.3 & 51.0 & 30.0 & 15.2 & 22.5 & 59.1 & 35.9 & 18.0 & 26.2 \\
\midrule
\multirow{4}{*}{\#Aspect-RM} & Direct & \underline{68.9} & \textbf{44.6} & \textbf{26.2} & \textbf{34.7} & \underline{73.0} & \underline{45.0} & 24.2 & 33.9 & 59.4 & 32.8 & 14.8 & 23.6 & \underline{71.4} & \underline{41.8} & \underline{21.1} & \underline{31.2} \\
 & CoT & 68.0 & 43.1 & \underline{25.6} & \underline{34.0} & \textbf{73.1} & 44.6 & \underline{25.0} & \textbf{34.9} & 59.5 & 31.5 & 15.3 & 23.8 & 68.1 & 36.0 & 17.7 & 27.7 \\
 & CoT-SC & \textbf{71.5} & \underline{43.6} & 23.7 & 32.6 & 71.8 & 41.9 & 21.5 & 31.4 & \underline{69.2} & \underline{37.4} & \underline{18.0} & \underline{27.8} & 71.2 & 40.1 & 20.0 & 30.1 \\
 & Self-Refine & 63.0 & 39.5 & 22.4 & 30.4 & 70.3 & \textbf{45.3} & \textbf{25.3} & \underline{34.7} & \textbf{70.2} & \textbf{41.2} & \textbf{20.4} & \textbf{30.7} & \textbf{71.7} & \textbf{43.2} & \textbf{21.5} & \textbf{31.7} \\
\bottomrule
\end{tabular}
}
\caption{Results (×100\%) on RW-ECT.}
\label{tab:rw-all}
\end{table*}

\subsection{Additional Models}
We additionally run experiments with two other reasoning models, o3-mini and Gemini-2.5-Flash-Lite, on the three datasets. 
As shown in Tables \ref{tab:ect-all-raw}, \ref{tab:covid-all-raw}, and \ref{tab:rw-all-raw}, \#Aspect-RM continues to achieve the highest scores, and o3-mini demonstrates the best performance under the Self-Refine strategy.

\subsection{Annotation Details}
\label{app:annotation}
\begin{figure*}[h]
    \centering
    \includegraphics[width=\linewidth]{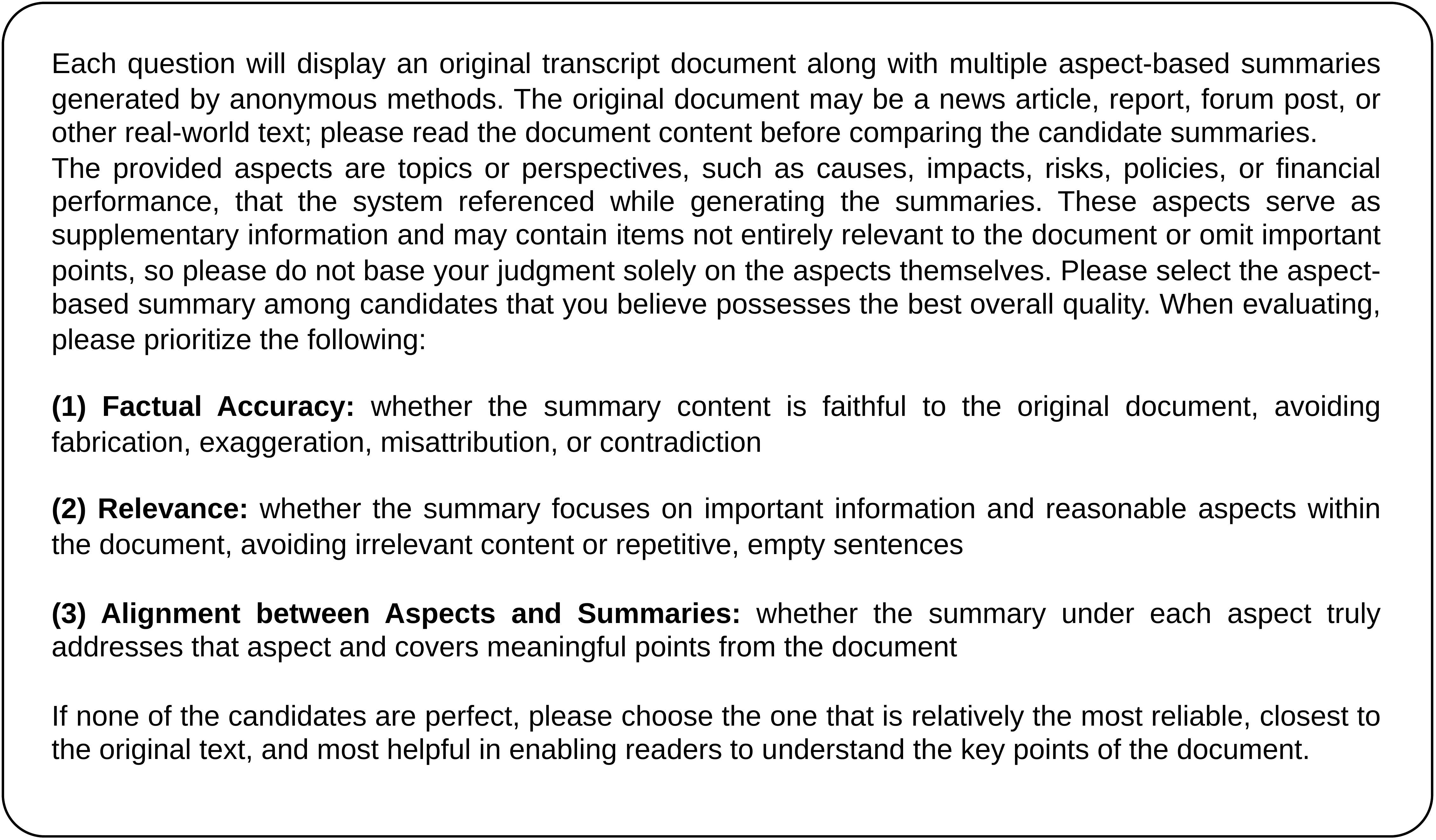}
    \caption{Annotation instructions provided to human annotators.}
    \label{fig:annotation_instruction}
\end{figure*}

We invite 12 college-based volunteers to act as human annotators.
Specifically, we evenly sample a total of 450 transcripts, generated aspects, and summaries from ECT and RW-ECT datasets across four prompting strategies on GPT-4o.
Then, the annotators are asked to select the aspects and summaries that reflect the transcripts the most.
The order of option aspects and summaries provided to annotators are randomly shuffled to mitigate order bias.
The instruction provided to the annotators is shown in Figure \ref{fig:annotation_instruction}.

\begin{table*}[ht]
\centering
\small
\caption{Results of two reasoning models on ECT. BERT refers to BERTScore, R-1 to ROUGE-1, R-2 to ROUGE-2, R-L to ROUGE-L, and \#AbsAspDiff is the absolute aspect difference.}
\setlength{\tabcolsep}{1mm}%
\begin{tabular}{c c c c c c c c c c c c}
\toprule
\multirow{2}{*}{} & \multirow{2}{*}{\textbf{Prompt}} 
& \multicolumn{5}{c}{\textbf{Gemini-2.5-Flash-Lite}} 
& \multicolumn{5}{c}{\textbf{o3-mini-2025-01-31}} \\
\cmidrule(lr){3-7} \cmidrule(lr){8-12}
& & \textbf{BERT} & \textbf{R-1} & \textbf{R-2} & \textbf{R-L} & \textbf{\#AbsAspDiff($\downarrow$)}
  & \textbf{BERT} & \textbf{R-1} & \textbf{R-2} & \textbf{R-L} & \textbf{\#AbsAspDiff($\downarrow$)} \\
\midrule
\multirow{4}{*}{Preliminary} 
& Direct & 0.630 & 0.452 & 0.275 & 0.356 & 1.963 & 0.565 & 0.332 & 0.145 & 0.238 & 1.539\\
& CoT          & 0.608 & 0.436 & 0.266 & 0.347 & 2.176 & 0.489 & 0.282 & 0.130 & 0.213 & 2.190\\
& CoT-SC       & 0.652 &  0.467 &  0.282 &  0.368 &  1.650 &  0.542 &  0.329 &  0.152 &  0.239 &  1.787\\
& Self-Refine  & 0.503 & 0.363 & 0.227 & 0.293 & 3.957 & 0.554 & 0.351 & 0.169 & 0.258 & 1.971\\
\midrule
\multirow{4}{*}{\#Aspect-LLM} 
& Direct       & 0.512 & 0.367 & 0.228 & 0.293 & 4.561 & 0.589 & 0.342 & 0.145 & 0.242 & 1.098\\
& CoT          & 0.496 & 0.353 & 0.213 & 0.281 & 4.831 & 0.582 & 0.320 & 0.139 & 0.238 & 1.249\\
& CoT-SC       & 0.504 & 0.356 & 0.211 & 0.280 & 4.246 & 0.604 & 0.345 & 0.148 & 0.247 & 1.087\\
& Self-Refine  & 0.496 & 0.357 & 0.223 & 0.287 & 4.536 & 0.619 & 0.379 & 0.172 & 0.272 & 1.088\\
\midrule
\multirow{4}{*}{\#Aspect-RM} 
& Direct       & \underline{0.741} & \underline{0.515} & \textbf{0.302} & \underline{0.398} & \underline{0.396} & 0.661 & 0.376 & 0.156 & 0.266 & 0.277 \\
& CoT          & 0.735 & 0.510 & \underline{0.298} & \textbf{0.399} & 0.525 & 0.656 & 0.358 & 0.155 & 0.266 & 0.465 \\
& CoT-SC       & \textbf{0.750} & \textbf{0.516} & 0.293 & 0.396 & \textbf{0.307} & \underline{0.680} & \underline{0.385} & \underline{0.163} & \underline{0.275} & \underline{0.257} \\
& Self-Refine  & 0.635 & 0.454 & 0.278 & 0.361 & 1.613 & \textbf{0.697} & \textbf{0.422} & \textbf{0.191} & \textbf{0.303} & \textbf{0.254 }\\
\bottomrule
\end{tabular}

\label{tab:ect-all-raw}
\end{table*}
\begin{table*}[ht]
\centering
\small
\caption{Results of two reasoning models on COVID-19-PC.}
\setlength{\tabcolsep}{1mm}%
\begin{tabular}{c c c c c c c c c c c c}
\toprule
\multirow{2}{*}{} & \multirow{2}{*}{\textbf{Prompt}} 
& \multicolumn{5}{c}{\textbf{Gemini-2.5-Flash-Lite}} 
& \multicolumn{5}{c}{\textbf{o3-mini-2025-01-31}} \\
\cmidrule(lr){3-7} \cmidrule(lr){8-12}
& & \textbf{BERT} & \textbf{R-1} & \textbf{R-2} & \textbf{R-L} & \textbf{\#AbsAspDiff($\downarrow$)}
  & \textbf{BERT} & \textbf{R-1} & \textbf{R-2} & \textbf{R-L} & \textbf{\#AbsAspDiff($\downarrow$)} \\
\midrule
\multirow{4}{*}{Preliminary} 
& Direct & 0.610 & 0.436 & 0.259 & 0.351 & 1.874 & 0.569 & 0.334 & 0.140 & 0.240 & 1.451\\
& CoT          & 0.533 & 0.373 & 0.220 & 0.303 & 2.721 & 0.489 & 0.270 & 0.112 & 0.202 & 1.933\\
& CoT-SC       & 0.586 & 0.421 & 0.252 & 0.341 & 2.325 & 0.556 & 0.337 & 0.148 & 0.245 & 1.519\\
& Self-Refine  & 0.506 & 0.368 & 0.225 & 0.304 & 3.448 & 0.557 & 0.351 & 0.159 & 0.258 & 1.834\\
\midrule
\multirow{4}{*}{\#Aspect-LLM} 
& Direct       & 0.512 & 0.367 & 0.221 & 0.299 & 3.601 & 0.604 & 0.347 & 0.139 & 0.246 & 0.882\\
& CoT          & 0.472 & 0.336 & 0.198 & 0.274 & 4.456 & 0.588 & 0.316 & 0.126 & 0.234 & 0.974\\
& CoT-SC       & 0.499 & 0.355 & 0.210 & 0.288 & 3.635 & 0.616 & 0.357 & 0.147 & 0.256 & 0.866\\
& Self-Refine  & 0.495 & 0.358 & 0.219 & 0.295 & 3.946 & 0.633 & 0.390 & 0.150 & 0.281 & 0.885\\
\midrule
\multirow{4}{*}{\#Aspect-RM} 
& Direct       & \underline{0.739} & \underline{0.511} & \textbf{0.293} & \underline{0.407} & \underline{0.275} & 0.680 & 0.385 & 0.151 & 0.273 & 0.053\\
& CoT          & 0.708 & 0.485 & 0.274 & 0.388 & 0.578 & 0.665 & 0.356 & 0.140 & 0.262 & 0.159\\
& CoT-SC       & \textbf{0.753} & \textbf{0.515} & \underline{0.291} & \textbf{0.410} & \textbf{0.128} & \underline{0.695} & \underline{0.399} & \underline{0.162} & \underline{0.285} & \textbf{0.041}\\
& Self-Refine  & 0.665 & 0.475 & 0.282 & 0.386 & 1.222 & \textbf{0.714} & \textbf{0.432} & \textbf{0.186} & \textbf{0.312} & \textbf{0.041}\\
\bottomrule
\end{tabular}

\label{tab:covid-all-raw}
\end{table*}
\begin{table*}[ht]
\centering
\small
\caption{Results of two reasoning models on RW-ECT.}
\setlength{\tabcolsep}{1mm}%
\begin{tabular}{c c c c c c c c c c c c}
\toprule
\multirow{2}{*}{} & \multirow{2}{*}{\textbf{Prompt}} 
& \multicolumn{5}{c}{\textbf{Gemini-2.5-Flash-Lite}} 
& \multicolumn{5}{c}{\textbf{o3-mini-2025-01-31}} \\
\cmidrule(lr){3-7} \cmidrule(lr){8-12}
& & \textbf{BERT} & \textbf{R-1} & \textbf{R-2} & \textbf{R-L} & \textbf{\#AbsAspDiff($\downarrow$)}
  & \textbf{BERT} & \textbf{R-1} & \textbf{R-2} & \textbf{R-L} & \textbf{\#AbsAspDiff($\downarrow$)} \\
\midrule
\multirow{4}{*}{Preliminary} 
& Direct & 0.581 & 0.396 & 0.237 & 0.308 & 3.047 & 0.549 & 0.302 & 0.123 & 0.211 & 1.592\\
& CoT          & 0.588 & 0.409 & 0.247 & 0.316 & 2.232 & 0.464 & 0.253 & 0.111 & 0.187 & 2.245\\
& CoT-SC       & 0.625 & 0.435 & 0.263 & 0.338 & 1.919 & 0.525 & 0.299 & 0.129 & 0.214 & 1.744\\
& Self-Refine  & 0.503 & 0.349 & 0.211 & 0.273 & 3.734 & 0.548 & 0.330 & 0.148 & 0.234 & 1.939\\
\midrule
\multirow{4}{*}{\#Aspect-LLM} 
& Direct       & 0.478 & 0.327 & 0.196 & 0.255 & 5.013 & 0.556 & 0.297 & 0.116 & 0.207 & 1.303\\
& CoT          & 0.466 & 0.318 & 0.191 & 0.251 & 5.457 & 0.555 & 0.288 & 0.121 & 0.213 & 1.414\\
& CoT-SC       & 0.477 & 0.319 & 0.186 & 0.248 & 4.532 & 0.561 & 0.299 & 0.119 & 0.211 & 1.370\\
& Self-Refine  & 0.475 & 0.325 & 0.197 & 0.258 & 5.144 & 0.583 & 0.332 & 0.139 & 0.231 & 1.303\\
\midrule
\multirow{4}{*}{\#Aspect-RM} 
& Direct       & \underline{0.728} & 0.481 & 0.271 & 0.366 & \underline{0.424} & 0.657 & 0.348 & 0.133 & 0.243 & 0.205\\
& CoT          & 0.722 & \underline{0.482} & \textbf{0.278} & \textbf{0.373} & 0.595 & 0.658 & 0.340 & \underline{0.140} & \underline{0.249} & 0.338\\
& CoT-SC       & \textbf{0.743} & \textbf{0.486} & \underline{0.272} & \underline{0.371} & \textbf{0.252} & \underline{0.670} & \underline{0.351} & 0.137 & 0.247 & \textbf{0.181}\\
& Self-Refine  & 0.673 & 0.458 & 0.271 & 0.356 & 1.175 & \textbf{0.690} & \textbf{0.391} & \textbf{0.167} & \textbf{0.274} & \textbf{0.181}\\
\bottomrule
\end{tabular}

\label{tab:rw-all-raw}
\end{table*}

\begin{table*}[ht]
\centering
\resizebox{\textwidth}{!}{%
\begin{tabular}{ll ccc ccc ccc ccc ccc ccc}
\toprule
\multirow{2}{*}{\textbf{Dataset}} & \multirow{2}{*}{\textbf{Method}} & \multicolumn{3}{c}{\textbf{Gemini-2.5-Flash}} & \multicolumn{3}{c}{\textbf{Gemma-3-12B}} & \multicolumn{3}{c}{\textbf{Gemma-3-27B}} & \multicolumn{3}{c}{\textbf{GPT-4o-mini}} & \multicolumn{3}{c}{\textbf{GPT-4o}} & \multicolumn{3}{c}{\textbf{O3-mini}} \\
\cmidrule(r){3-5} \cmidrule(r){6-8} \cmidrule(r){9-11} \cmidrule(r){12-14} \cmidrule(r){15-17} \cmidrule(r){18-20}
& & \textbf{AS} & \textbf{BS($\uparrow$)} & \textbf{AAD($\downarrow$)} & \textbf{AS} & \textbf{BS($\uparrow$)} & \textbf{AAD($\downarrow$)} & \textbf{AS} & \textbf{BS($\uparrow$)} & \textbf{AAD($\downarrow$)} & \textbf{AS} & \textbf{BS($\uparrow$)} & \textbf{AAD($\downarrow$)} & \textbf{AS} & \textbf{BS($\uparrow$)} & \textbf{AAD($\downarrow$)} & \textbf{AS} & \textbf{BS($\uparrow$)} & \textbf{AAD($\downarrow$)} \\
\midrule
\multirow{3}{*}{ECT} & Preliminary & 3.36 & 0.50 & 3.96 & 3.20 & 0.57 & 2.42 & 2.30 & 0.52 & 2.65 & 11.16 & 0.53 & 3.15 & 2.75 & 0.57 & 2.21 & 3.34 & 0.55 & 1.97 \\
& \#Aspect-LLM & 3.39 & 0.50 & 4.54 & 3.14 & 0.56 & 2.53 & 2.30 & 0.53 & 2.39 & 3.42 & 0.53 & 2.92 & 3.24 & 0.55 & 2.10 & 3.90 & 0.62 & 1.09 \\
& \#Aspect-RM & 3.38 & 0.63 & 1.61 & 5.09 & 0.65 & 1.31 & 3.59 & 0.71 & 0.78 & 7.15 & 0.71 & 0.42 & 4.17 & 0.72 & 0.29 & 3.86 & 0.70 & 0.25 \\
\midrule
\multirow{3}{*}{COVID-19-PC} & Preliminary & 3.32 & 0.51 & 3.45 & 3.21 & 0.60 & 1.90 & 2.52 & 0.52 & 2.49 & 3.58 & 0.53 & 2.20 & 3.36 & 0.56 & 2.03 & 3.43 & 0.56 & 1.83 \\
& \#Aspect-LLM & 3.67 & 0.50 & 3.95 & 5.86 & 0.54 & 2.93 & 3.65 & 0.60 & 1.80 & 5.44 & 0.54 & 2.48 & 4.00 & 0.57 & 1.93 & 3.92 & 0.63 & 0.89 \\
& \#Aspect-RM & 3.45 & 0.67 & 1.22 & 6.15 & 0.67 & 1.32 & 3.56 & 0.72 & 0.56 & 5.08 & 0.71 & 0.25 & 3.85 & 0.73 & 0.10 & 3.86 & 0.71 & 0.04 \\
\midrule
\multirow{3}{*}{RW-ECT} & Preliminary & 3.25 & 0.50 & 3.73 & 3.24 & 0.55 & 2.56 & 2.32 & 0.51 & 2.62 & 3.59 & 0.51 & 2.37 & 3.27 & 0.55 & 2.17 & 3.38 & 0.55 & 1.94 \\
& \#Aspect-LLM & 4.01 & 0.48 & 5.14 & 5.03 & 0.45 & 4.98 & 3.39 & 0.58 & 2.11 & 7.10 & 0.51 & 3.24 & 3.95 & 0.59 & 1.59 & 3.95 & 0.58 & 1.30 \\
& \#Aspect-RM & 3.53 & 0.67 & 1.18 & 6.53 & 0.63 & 1.82 & 3.43 & 0.70 & 0.73 & 6.91 & 0.70 & 0.33 & 3.90 & 0.72 & 0.24 & 3.85 & 0.69 & 0.18 \\
\bottomrule
\end{tabular}%
}
\caption{Results of Self-Refine on the three dataset, showing average thinking steps (AS), BERTScore (BS), and \#AbsAspDiff (AAD).}
\label{tab:combined_self_refine}
\end{table*}

\section{Default Aspect Sets, Dataset \& Prompts}
\label{app:Dataset_Detailed}
\begin{figure*}[ht]
    \centering
    \begin{subfigure}[b]{0.45\textwidth}
        \includegraphics[width=\linewidth]{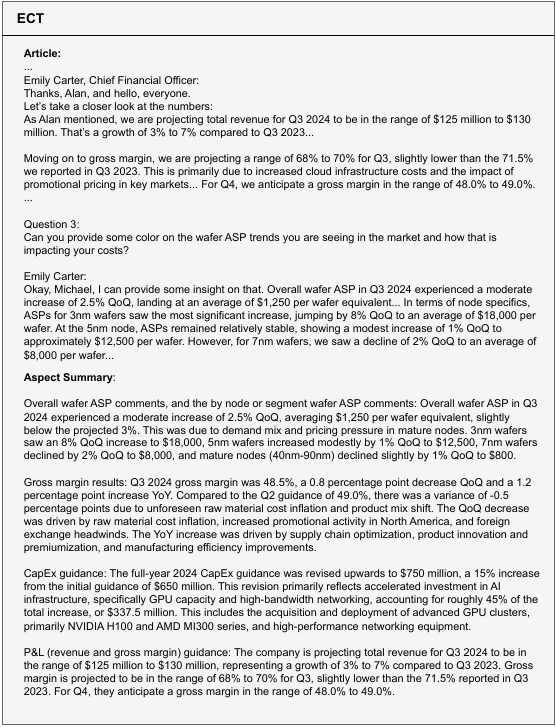}
        \caption{ECT}
        \label{fig:ect-dataset}
    \end{subfigure}
    \hfill
    \begin{subfigure}[b]{0.45\textwidth}
        \centering
        \includegraphics[width=\linewidth]{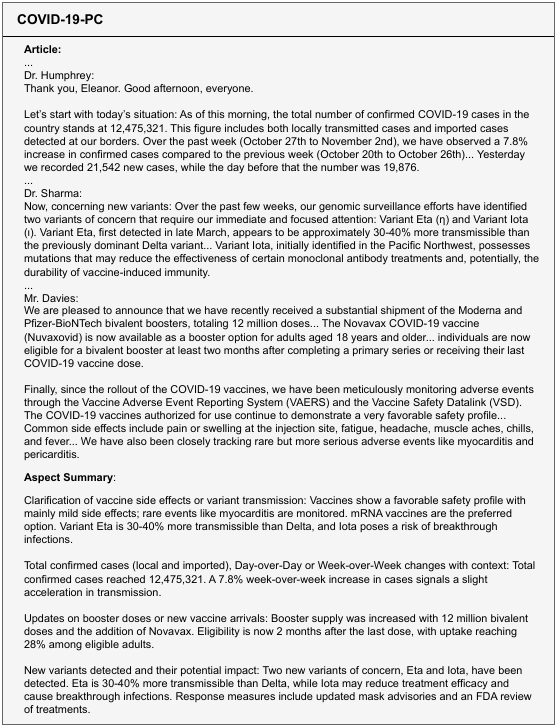}
        \caption{COVID-19-PC}
        \label{fig:covid-dataset}
    \end{subfigure}
    \caption{Examples of the two synthetic datasets.}
\end{figure*}

Furthermore, we provide the default aspect example in Table \ref{tab:default_aspect_example}, and data example in Figure \ref{fig:ect-dataset} and \ref{fig:covid-dataset}.

\section{Use of AI assistants}
We use AI to help our coding and grammar correction in our paper writing.
We do not use AI to help us in related paper surveys.

\end{document}